\documentclass{article}

\usepackage{xcolor} 
\usepackage[final]{corl_2026}
\usepackage[T1]{fontenc}
\usepackage{graphicx}
\usepackage{booktabs,multirow,tabularx,colortbl}
\usepackage{algorithm,algpseudocode}
\usepackage{amsmath,amssymb,bbm}
\usepackage{microtype}
\usepackage{placeins}
\usepackage{subcaption}
\usepackage[accsupp]{axessibility}
\ifdefined\pdfsuppressptexinfo\pdfsuppressptexinfo=-1\fi

\definecolor{mplC0}{HTML}{1F77B4}
\definecolor{mplC1}{HTML}{FF7F0E}
\definecolor{mplC2}{HTML}{2CA02C}
\definecolor{mplC3}{HTML}{D62728}

\makeatletter
\renewcommand{\@noticestring}{CoRL 2026 Workshop on Spatial Intelligence and Structured Reasoning for Embodied Navigation (SISREN).}
\makeatother

\title{VLD: Visual Learned Distance as an Interface\\for Reinforcement Learning Navigation}
\author{Lazar Milikic$^{1}$ \qquad Manthan Patel$^{2}$ \qquad Jonas Frey$^{3}$ \\
  \normalfont $^{1}$EPFL \qquad $^{2}$ETH Zurich \\
  \normalfont $^{3}$University of California, Berkeley \\[0.4em]
  \normalfont\small\textbf{Project:} \url{https://leggedrobotics.github.io/rl_distance_nav/}
}
\hypersetup{
  pdftitle={VLD: Visual Learned Distance as an Interface for Reinforcement Learning Navigation},
  pdfauthor={Lazar Milikic, Manthan Patel, Jonas Frey},
  pdfsubject={SISREN Workshop at CoRL 2026}
}

\begin{document}
\maketitle
\begin{abstract}
Training end-to-end policies from image data to directly predict navigation actions for robotic systems has proven inherently difficult. Existing approaches often suffer from either the sim-to-real gap during policy transfer or limited action-labeled training data. To address this problem, we introduce Visual Learned Distance (VLD), a scalable framework for image-goal navigation that decouples perception learning from policy learning. We first train a self-supervised distance-to-goal predictor on about 3,000 hours of internet-scale and simulated video, providing a distance signal that a reinforcement learning (RL) policy can minimize. The RL policy is trained separately in simulation using privileged geometric distances perturbed by GeoNoise, a learned noise model that reproduces the viewpoint- and overlap-dependent errors of the trained distance predictor. At deployment, the policy consumes VLD predictions zero-shot, without retraining, inheriting ``where to go'' from large-scale visual training while retaining the robust navigation behaviors learned in simulation. This error modeling is key to the transfer: with VLD inputs, GeoNoise-trained policies reach $58.7\%$ success on HM3D and $76.2\%$ on Gibson, whereas policies trained with generic temporally correlated noise collapse to $4.9\%$ and $11.5\%$. On a physical robot, the VLD-driven policy succeeds in $88\%$ of trials, compared with $60\%$ for a strong end-to-end image-goal baseline that outperforms it in simulation. Our decoupled design thus provides an alternative and scalable path toward reliable navigation policies that transfer from simulation to the real world.
\end{abstract}

\section{Introduction}
\label{sec:introduction}

Scalable training on diverse, real-world(-like), internet-scale datasets has emerged as one of the most promising approaches for learning broadly capable navigation policies. Recent advances in embodied AI show that policies trained on large-scale multimodal data exhibit robustness and generalization that narrow, task-specific approaches fail to achieve~\cite{NaVILA}, highlighting the importance of navigation research under conditions that mirror the real world rather than controlled lab settings~\cite{city,roboticsChallnages}.

Imitation learning from expert demonstrations offers a straightforward pathway for training navigation agents~\cite{ChauffeurNet}, but expert-labeled actions---typically future 2D or 2.5D waypoints---are expensive to collect, so these pipelines remain constrained by the scarcity of annotated trajectories~\cite{navitrace}. Large-scale efforts such as GNM~\cite{gnm}, ViNT~\cite{vint}, and NoMaD~\cite{nomad} collect on the order of 200 hours of demonstrations across robotic embodiments; nevertheless, these datasets exhibit a strong \emph{straight-walking bias}, raising doubts about whether annotation-heavy pipelines can scale to broadly generalizable navigation.

Reinforcement learning (RL) instead enables agents to develop robustness through interaction and exploration. Due to its sample inefficiency in the real world, however, RL for navigation depends heavily on simulators, and collecting diverse environment assets and simulating photorealistic images remain open challenges that hinder the transfer of visual navigation policies from simulation to reality~\cite{10669238,city,habitat19iccv,roboticsChallnages,vr-robo}.

\begin{figure*}[t]
  \centering
  \includegraphics[width=0.85\linewidth]{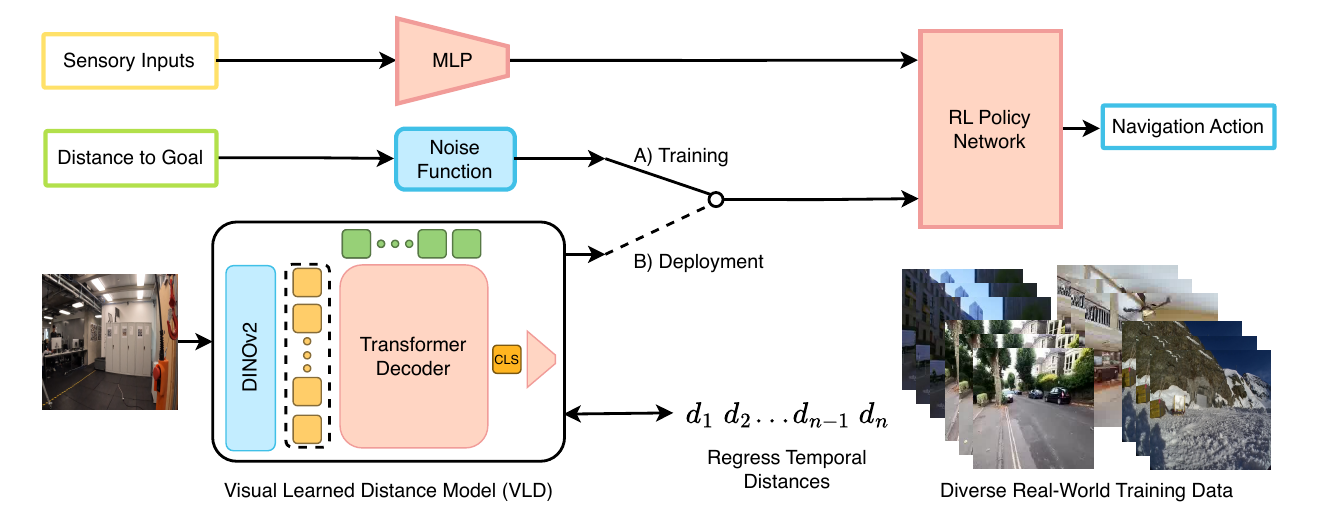}
  \caption{\textbf{Overview of our framework.} (A) Training: we separately train a temporal Visual Learned Distance (VLD) function on diverse real-world and synthetic video, and an RL navigation policy in simulation that receives the privileged distance-to-goal perturbed by GeoNoise, a learned model of VLD's errors. (B) Deployment: the trained policy consumes VLD predictions in place of the noisy privileged distance, without retraining, to navigate in simulated and real-world environments.}
  \label{fig:overview}
\end{figure*}

In this work, we combine the strengths of both paradigms while avoiding their annotation and sim-to-real bottlenecks by decoupling navigation into two stages (Figure~\ref{fig:overview}): (i) training a scalable, self-supervised distance-to-goal predictor, \emph{Visual Learned Distance} (VLD), on internet-scale video, and (ii) training RL policies in simulation that consume this distance signal directly. Unlike expert-labeled trajectories, the distance signal---the number of temporal steps from the current view to a goal observation---can be obtained from sequential video without annotation, making it both scalable and domain-agnostic. We curate about 3,000 hours of diverse trajectories from synthetic environments~\cite{habitat19iccv}, internet-scale walking videos~\cite{city,ego4d}, and embodiment datasets~\cite{scanD,GTDataset}. Large-scale video learning thus provides the perception of \emph{where} to navigate, while RL in simulation learns \emph{how} to navigate from a geometric goal signal that transfers robustly from simulation to the real world~\cite{lee2024learning}.

The key challenge of this decoupling is the transfer step itself. A policy trained on exact simulator distances learns to trust a signal that no learned predictor can provide: the errors of VLD are not random but structured---its predictions change as the robot turns, and become more uncertain when the goal is out of view. We therefore train the policy with \emph{GeoNoise}, a lightweight learned model that maps privileged geometry, namely the visual overlap and relative pose between the current view and the goal, to the distribution of VLD predictions. The policy experiences predictor-like distances throughout RL training without running VLD in the simulation loop, and at deployment VLD replaces GeoNoise \emph{zero-shot}, without retraining. Success with a noisy privileged input does not predict this transfer: policies trained with generic temporally correlated noise succeed \emph{in simulation} more often than GeoNoise-trained ones before substitution, yet collapse after it ($4.9\%$ versus $58.7\%$ SR on HM3D). The resulting interface lets perception and control scale independently: perception improves with more diverse video, and control with more simulation. On a physical robot, the VLD-driven policy succeeds in $88\%$ of trials versus $60\%$ for FGPrompt-EF~\cite{fgprompt2023}, reversing the ranking observed in simulation.

Our main contributions are: (i) a scalable framework for training \textbf{visual distance-to-goal predictors} on internet-scale video, with \emph{ordinal consistency} as a direct evaluation of temporal distance functions, under which VLD outperforms prior designs such as ViNT~\cite{vint} and VIP~\cite{vip}; (ii) \textbf{GeoNoise}, an overlap-conditioned model of predictor errors that, unlike generic noise, enables zero-shot substitution of a learned predictor into a policy trained in simulation; and (iii) a \textbf{navigation framework} that decouples RL training in simulation from perception learning, validated in simulation and in real-world deployment, where it outperforms a strong end-to-end image-goal baseline.

\section{Related Work}\label{sec:related-work}

\textbf{Temporal distance functions.}
Temporal distance has evolved from graph-based heuristics to scalable self-supervised learning. ViNG regresses temporal steps between observation pairs with negative mining and plans over an experience graph~\cite{VING}, but lacks closed-loop RL policies for robustness. Building on this idea, ViNT and NoMaD apply Transformer encoders over observation and goal embeddings, supervising both distance and short-horizon actions~\cite{vint,nomad}, but treat distance primarily as an auxiliary to the policy rather than evaluating the estimator in isolation. VIP reframes representation learning as goal-conditioned value prediction, where Euclidean distance in latent space acts as a reward~\cite{vip}, and quasimetric RL formalizes multi-goal optimal values as quasimetric distances~\cite{qrl,qrl2}. Yet most evaluations emphasize downstream success rather than direct validation of the distance function itself. We adopt pairwise supervision with negative mining, but design a Transformer \emph{decoder} that queries distance from the goal, train it with an inlier--outlier Gaussian mixture NLL~\cite{gaus-nill2}, and evaluate it directly through ordinal consistency.

\textbf{Navigation policies.}
Goal-conditioned RL trains policies conditioned on state or image goals~\cite{GCRL}. While point-goal navigation is largely solved~\cite{point-goal}, image-goal navigation remains challenging due to its reliance on simulators and the resulting sim-to-real gap~\cite{vr-robo,sim2real}. End-to-end image-goal RL methods such as FGPrompt~\cite{fgprompt2023} learn goal-conditioned visual features jointly with control and achieve strong results in simulation, but must carry their visual processing over from simulated to real images. Vision--language--action models~\cite{NaVILA,openVLA} couple perception and control in a single model at a high computational cost, and imitation learning~\cite{drivingIL,visgoalnav,gnm,vint,nomad,city} remains limited by annotation costs. Our approach explicitly decouples perception from control: goal perception is learned by self-supervised distance learning on video without action labels, and control is learned with RL in simulation from a scalar goal signal, reducing the dependence on simulator photorealism~\cite{escontrela2025gaussgym}.

\textbf{Privileged training and perception noise.}
Privileged-information and distillation approaches transfer knowledge available only in simulation to deployable policies~\cite{anymalPriv,RMA,fu2022learningdeepsensorimotorpolicies}, often injecting hand-designed noise into privileged observations to make policies robust to imperfect perception~\cite{anymalPriv,jin2023resilientleggedlocalnavigation}. In contrast, we directly substitute a raw privileged value with a learned predictor at deployment and, to make this substitution work, learn the error distribution of that specific predictor as a function of visual overlap and relative pose.

\section{Method}
\label{sec:method}

\subsection{Visual Learned Distance}
\label{sec:temporal_dist}
We define the temporal distance function as a mapping $\mathcal{T}_\theta(o,g)\rightarrow(\hat t,\hat c)$ from the current egocentric RGB observation $o$ and a goal image $g$ to the predicted number of steps $\hat t$ required to reach the goal, together with a confidence score $\hat c$. Unlike action-labeled supervision, temporal distance is naturally available from sequential video or trajectory data: two frames $o_i$ and $o_j$ of the same trajectory, with $i\le j$, are $td=j-i$ steps apart. This makes temporal distance a scalable, annotation-free signal that can be mined from large-scale video datasets.

\textbf{Evaluation methodology.}
\label{sec:evaluation_method}
Our goal is not to obtain perfectly accurate predictions of temporal distance, which is inherently ambiguous and often impossible to infer exactly from a single view (e.g., the distance from a bedroom to a kitchen varies drastically across apartments). Instead, what matters is that the learned function provides a \emph{useful ordering}: it should reliably distinguish between near and far goals, even if the absolute values are imprecise. We evaluate this property with \emph{ordinal consistency}, which measures whether predicted distances decrease as the agent approaches the goal and increase as it moves away. We quantify it with \emph{Kendall's~$\tau$}~\cite{Kendall1938-ik}, a rank correlation coefficient in $[-1,1]$, where $\tau=1$ indicates perfect agreement, $\tau=-1$ complete disagreement, and $\tau=0$ random ordering. It is fully scale-invariant, in line with our focus on relative rather than absolute accuracy.

\begin{figure*}[t]
\centering
\includegraphics[width=0.88\linewidth]{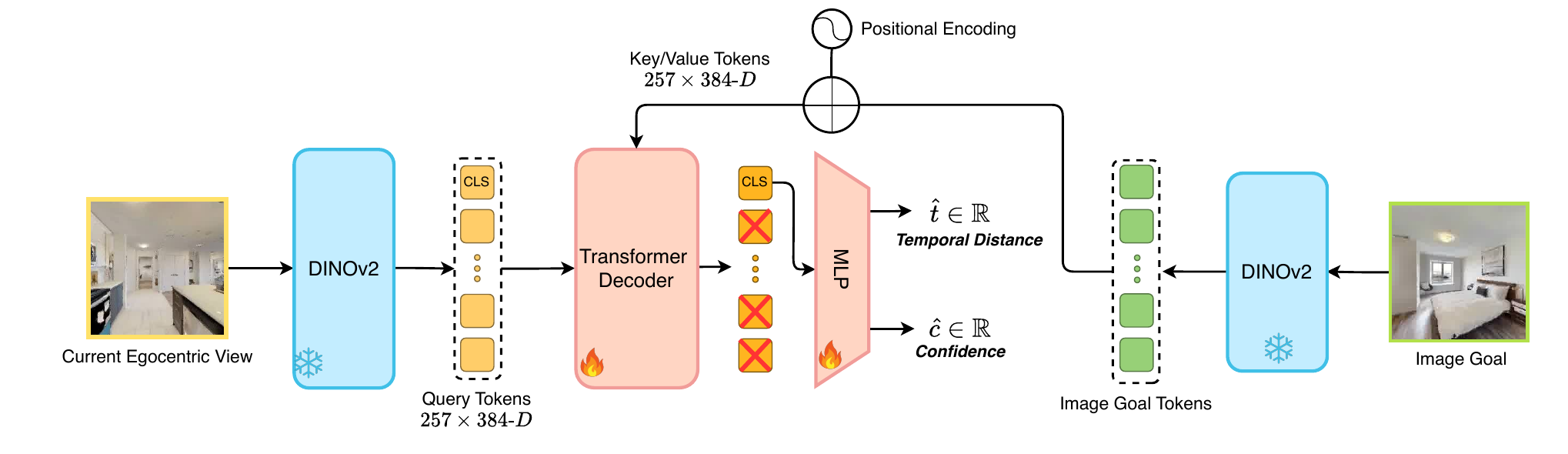}
\caption{\textbf{Visual Learned Distance (VLD) architecture.} A shared frozen DINOv2 encoder turns the current view into query tokens and the goal image into key/value memory tokens. A Transformer decoder attends from the observation queries to the goal tokens, and MLP heads on the CLS output predict the temporal distance $\hat t$ and a confidence score $\hat c$.}
\label{fig:vld}
\end{figure*}

\textbf{Model architecture.}
\label{sec:model_arch}
The current observation and the goal image are encoded with a shared frozen DINOv2-Small backbone~\cite{dino} (Figure~\ref{fig:vld}); each $224\times224$ image yields $257$ tokens of dimension $384$, including a CLS token. Observation tokens form the queries of an $L$-layer Transformer decoder, and goal tokens, with positional embeddings, form its key/value memory. From the decoder output $Z$ at the CLS token, two MLP heads predict the temporal distance $\hat{t} = \mathrm{ReLU}(\mathrm{MLP}_t(Z_{\text{CLS}}))\ge 0$ and the confidence $\hat{c} = \sigma(\mathrm{MLP}_c(Z_{\text{CLS}}))\in[0,1]$.

\textbf{Training procedure.}
\label{sec:model_loss}
We frame distance prediction as a self-supervised regression problem over triplets $(o_i,o_j,td=j-i)$ sampled along trajectories with $td\le td_{\max}$. Following ViNG~\cite{VING}, we also pair observations from different trajectories and assign them the maximum distance $td_{\max}$ (negative mining; Appendix~\ref{appdx:train_algo}). Because exact distance prediction is inherently noisy and often ambiguous, the confidence lets the model down-weight uncertain pairs (e.g., long-horizon pairs with little visual overlap) while still exploiting reliable near-goal predictions. Concretely, we model the likelihood of the ground-truth distance $td_n$ as an \emph{inlier--outlier Gaussian mixture}~\cite{gaus-nill2}: with probability $\hat c_n$, the error is drawn from a tight ``reliable'' Gaussian with variance $\sigma_R^2$, and with probability $1-\hat c_n$ from a broad ``outlier'' Gaussian with variance $\sigma_O^2$, where $\sigma_R\ll\sigma_O$. The training objective is the negative log-likelihood (NLL) over a batch of $N$ samples:
\begin{equation}
\mathcal{L} = -\frac{1}{N}\sum_{n=1}^{N}\log\Bigl[\,\hat c_n\,\mathcal{N}\!\left(td_n;\hat t_n,\sigma_R^2\right) + (1-\hat c_n)\,\mathcal{N}\!\left(td_n;\hat t_n,\sigma_O^2\right)\Bigr],
\label{eq:loss}
\end{equation}
where $\mathcal{N}(x;\mu,\sigma^2)$ denotes the Gaussian density. Compared with mean squared error, this objective improves ordinal consistency in five of six Habitat settings and matches it in the sixth (Appendix~\ref{appdx:objectives}).

\textbf{Training data.}
We combine three complementary sources totaling about 3,000 hours: (i) synthetic trajectories generated in Habitat~\cite{habitat19iccv} on HM3D scenes~\cite{hm3d}, where a shortest-path follower walks to an annotated goal object and finishes facing it; (ii) ``in-the-wild'' walking videos from YouTube~\cite{city} and Ego4D~\cite{ego4d}; and (iii) embodiment datasets collected on real robots~\cite{scanD,GTDataset,city}. We resample all sources so that consecutive frames are approximately $0.25$\,m apart, which makes the supervision units comparable across sources (details in Appendix~\ref{appdx:data_collection}).

\subsection{Navigation Policy with GeoNoise}
\label{sec:nav_policy}

We formulate navigation as a variant of the \emph{point-goal navigation} task: the agent is placed in a 3D environment with an unknown map and must reach a goal location specified only by its distance from the agent's current position. During training, this distance-to-goal signal comes from the simulator; at deployment, it is replaced by VLD predictions computed from the current camera image and the goal image. This setup lets the RL agent learn navigation behaviors independently of simulator photorealism for goal recognition, while grounding its decisions in a scalar signal that can also be inferred from real-world images. Since the goal enters only through this scalar, the policy must reason about spatial structure implicitly, through how the distance evolves as the agent moves and turns.

\textbf{Policy network.}
\label{sec:policy_net}
The policy receives (i) \emph{sensory information}---lightweight obstacle-awareness features from a frozen pretrained EfficientNet-B0~\cite{effNet} applied to the RGB observation, GPS displacement, and proprioception---and (ii) the \emph{distance-to-goal}. Distance inputs share a common scale: privileged distances are divided by the initial distance $d_0$ of the episode, so that the input decreases from $1$ at the start to $0$ at the goal, and VLD predictions are divided by the training horizon $td_{\max}=100$. Inputs are fed into an LSTM core that maintains temporal memory, followed by a policy head that outputs discrete navigation actions and a value head used for PPO optimization~\cite{ppo}. The reward extends the Zero-Exploration Reward~\cite{zer} with a visual-overlap term that encourages the agent to orient toward the goal (Appendix~\ref{appdx:reward}).

\textbf{GeoNoise: training with predictor-like distances.}
\label{sec:noise_f}
The errors of VLD are structured: the simulator provides geometric distance while VLD estimates temporal distance from images, so the two are only partially correlated, and the dominant source of prediction error is the degree of visual overlap between the agent's view and the goal. To expose the policy to this structure, we learn a noise model, GeoNoise, that maps privileged geometry to the distribution of VLD outputs. For each observation--goal pair, we extract a 13-dimensional feature vector: (i) the \emph{projection success ratio} (PSR), the fraction of goal-image pixels that project into the current view with depth-consistent matches, as a measure of visual overlap (Appendix~\ref{sec:psr_appendix}); (ii) the relative camera rotation $R_{\text{rel}}\in SO(3)$, flattened to 9 values~\cite{geist2024learning}; and (iii) the relative translation $t_{\text{rel}}\in\mathbb{R}^3$, compressed with $\mathrm{SymLog}(x)=\mathrm{sign}(x)\log(1+\alpha|x|)$. A lightweight MLP maps these features to categorical distributions over distance and confidence bins and is trained with a cross-entropy loss on VLD predictions for HM3D image pairs. During RL training, we sample a bin $b$ from the predicted distribution and draw the final value from a Gaussian centered at the bin midpoint, $\hat z\sim\mathcal{N}\bigl(\tfrac{l_b+u_b}{2},(\tfrac{u_b-l_b}{6})^2\bigr)$, where $[l_b,u_b]$ is the bin interval, which yields smooth samples that respect the predicted uncertainty. The resulting noisy distance reproduces the error patterns of VLD: larger errors when the goal is out of view or behind the agent, and tight estimates when visual alignment is strong (Appendix~\ref{appdx:ou_noise_abl}).

GeoNoise has two properties that make the decoupling practical. First, it is cheap: a small MLP over geometric features replaces a 33M-parameter vision model in the simulation loop, so RL training never runs VLD. Second, it turns deployment into a \emph{zero-shot substitution}: since the policy is trained on a signal with the same structure as the one it later receives, VLD replaces GeoNoise without any policy fine-tuning. The main policy uses only the distance; GeoNoise also models the confidence, which we evaluate as an optional policy input in Section~\ref{sec:rl_application}.

\section{Experiments}
\label{sec:experiments}

\subsection{Distance Function Evaluation}
\label{sec:dist_eval}

\textbf{Baselines.}
\label{sec:baselines}
We compare VLD against two representative families of image-goal distance functions. The first is ViNT~\cite{vint}, a strong baseline due to its training on diverse real-world robot data and a parameter count comparable to ours ($\sim$31M); we evaluate the released model and a variant fine-tuned on our data (ViNT-Tuned). The second is VIP~\cite{vip}, which estimates distances as $\ell_2$ norms between image embeddings without cross-attention. The original VIP model uses a ResNet-50 backbone~\cite{resnet} and was developed for manipulation; for a fair comparison, we evaluate the released model, a version retrained on our navigation data (VIP-Nav), and a variant with a frozen DINOv2 encoder (VIP-DINOv2). Architectures and training details are given in Appendix~\ref{appdx:arch}.

\textbf{Data and metric.}
\label{sec:ordinal_eval}
We train VLD in three configurations: (i) synthetic Habitat data only, (ii) real-world in-the-wild and embodiment data only, and (iii) all sources. All models are evaluated on held-out validation trajectories: trajectories generated in the standard HM3D and Gibson validation scenes (Gibson~\cite{Gibson} is never used for VLD training) and held-out real-world trajectories. The final frame of each trajectory serves as the goal, so a perfect distance function produces a monotonically decreasing sequence of predictions. We compute Kendall's~$\tau$ on trajectory segments with maximum start-to-goal separations of 20, 50, and 100 steps: short horizons contain strong visual overlap with the goal, while long horizons increasingly include views in which the goal lies outside the field of view.

\begin{table*}[t]
  \centering
  \small
  \setlength{\tabcolsep}{4.5pt}
  \caption{\textbf{Ordinal consistency} (Kendall's~$\tau$, $\uparrow$) on held-out Habitat and real-world trajectories at horizons of 20, 50, and 100 steps. VLD is trained on synthetic data, real-world data, or all sources.}
  \label{tab:ordinal_consistency_all}
  \begin{tabular}{@{}lcccccc@{\hspace{8pt}}cccc@{}}
    \toprule
    \multirow{3}{*}{\textbf{Model}} &
    \multicolumn{6}{c}{\textbf{Habitat}} &
    \multicolumn{4}{c}{\textbf{Real-world}} \\
    \cmidrule(lr){2-7} \cmidrule(l){8-11}
    &
    \multicolumn{3}{c}{\textbf{HM3D}} &
    \multicolumn{3}{c}{\textbf{Gibson}} &
    \multicolumn{2}{c}{\textbf{In-the-wild}} &
    \multicolumn{2}{c}{\textbf{Embodiment}} \\
    \cmidrule(lr){2-4} \cmidrule(lr){5-7} \cmidrule(lr){8-9} \cmidrule(l){10-11}
    & 20 & 50 & 100 & 20 & 50 & 100 & 50 & 100 & 50 & 100 \\
    \midrule
    ViNT             & 0.40 & 0.35 & 0.28 & 0.53 & 0.41 & 0.39 & 0.35 & 0.25 & 0.41 & 0.32 \\
    ViNT-Tuned       & 0.67 & 0.56 & 0.48 & 0.78 & 0.65 & 0.64 & 0.40 & 0.29 & 0.48 & 0.37 \\
    VIP              & 0.29 & 0.19 & 0.17 & 0.39 & 0.30 & 0.29 & 0.28 & 0.20 & 0.43 & 0.33 \\
    VIP-Nav          & 0.55 & 0.45 & 0.38 & 0.65 & 0.51 & 0.50 & 0.32 & 0.23 & 0.46 & 0.39 \\
    VIP-DINOv2       & 0.42 & 0.45 & 0.40 & 0.62 & 0.52 & 0.51 & 0.42 & 0.30 & 0.52 & 0.44 \\
    \midrule
    VLD (synthetic)  & 0.81 & \textbf{0.71} & \textbf{0.62} & \textbf{0.84} & \textbf{0.74} & \textbf{0.71} & 0.44 & 0.31 & 0.58 & 0.48 \\
    VLD (real-world) & 0.10 & 0.07 & 0.07 & 0.12 & 0.10 & 0.09 & 0.23 & 0.18 & 0.16 & 0.14 \\
    VLD (all)        & \textbf{0.82} & 0.70 & 0.61 & \textbf{0.84} & 0.73 & \textbf{0.71} & \textbf{0.69} & \textbf{0.61} & \textbf{0.73} & \textbf{0.63} \\
    \bottomrule
  \end{tabular}
\end{table*}

\begin{figure}[!htbp]
\centering
\begin{subfigure}{0.97\linewidth}
  \includegraphics[width=\linewidth]{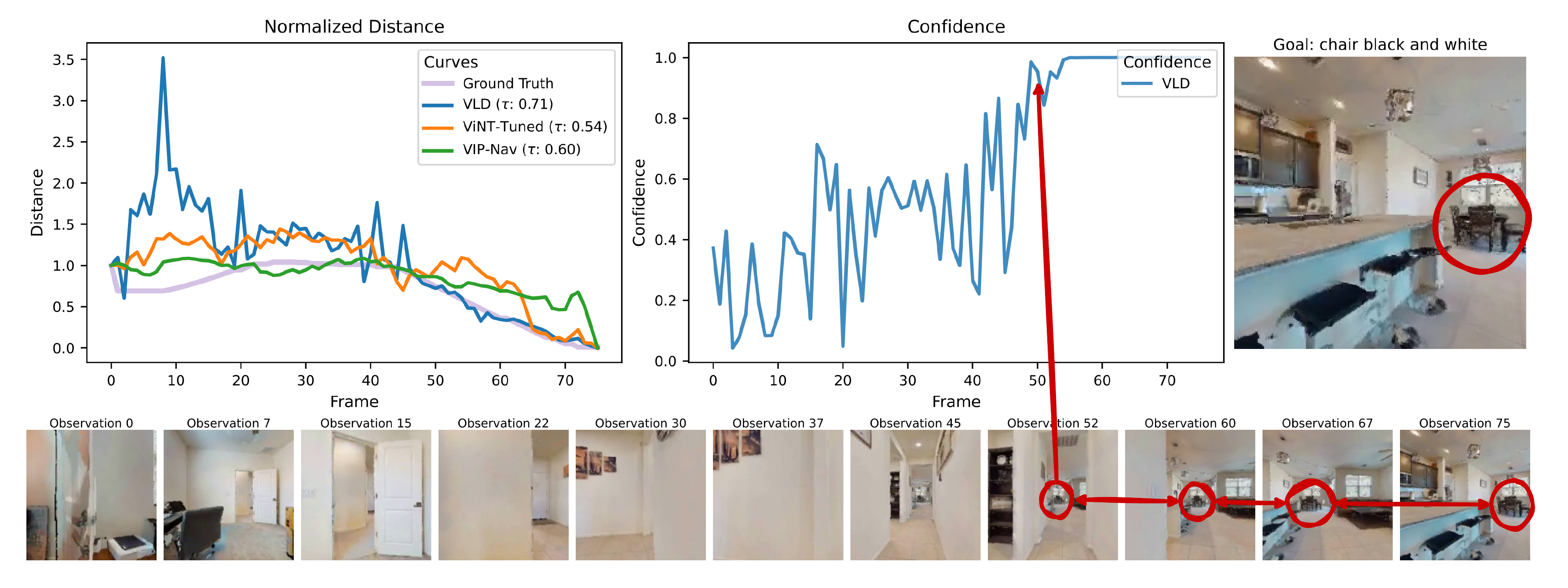}
  \caption{Image-goal Habitat trajectory: the chair goal becomes visible as the camera enters the room.}
  \label{fig:goal_cues_habitat}
\end{subfigure}
\par\smallskip
\begin{subfigure}{0.97\linewidth}
  \includegraphics[width=\linewidth]{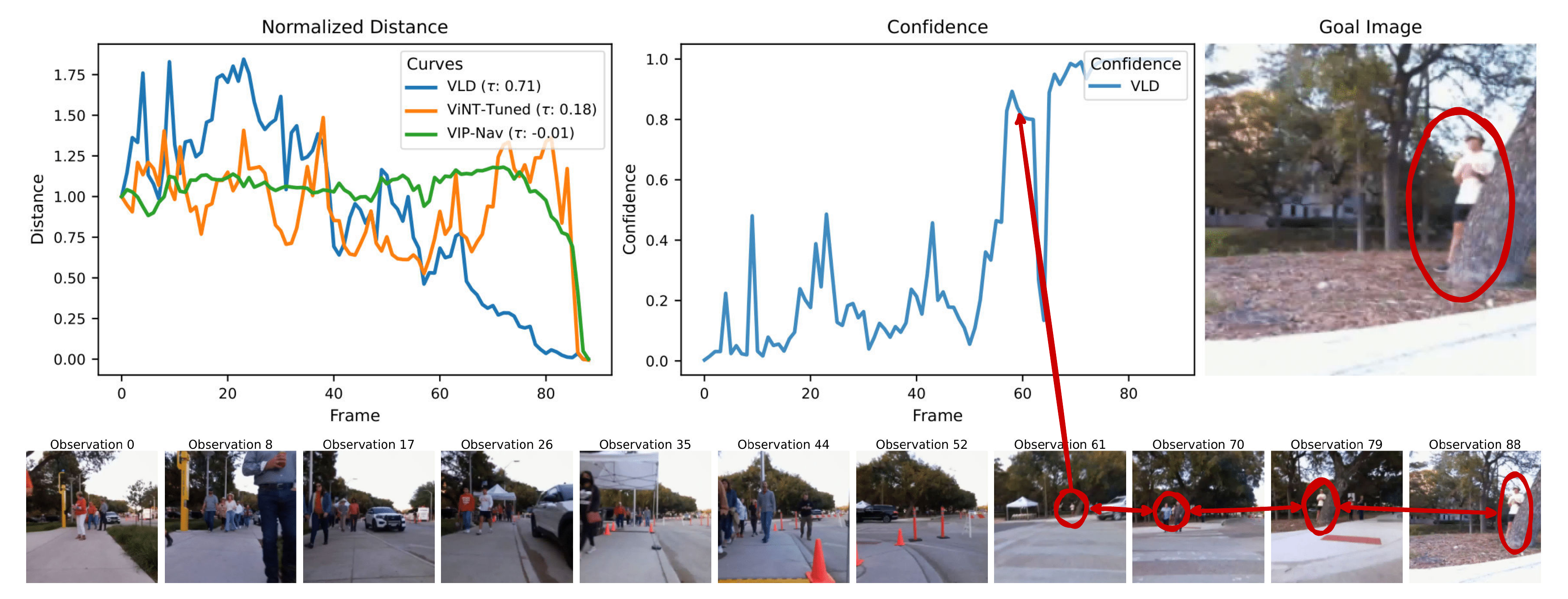}
  \caption{Image-goal real-world video: the pedestrian in the goal image appears with a small visual footprint.}
  \label{fig:goal_cues_real}
\end{subfigure}
\caption{\textbf{Goal cues and distance predictions.} Each observation is evaluated independently against the final-frame goal image. Left: normalized predicted distances and Kendall's~$\tau$; middle: VLD confidence; right: the goal image; below: observed frames. Red annotations link emerging goal cues to increases in confidence. Confidence is auxiliary; only distance supplies goal information to the main navigation policy.}
\label{fig:goal_cues}
\end{figure}

\textbf{Results.}
Table~\ref{tab:ordinal_consistency_all} reports ordinal consistency. On Habitat data, VLD models trained with synthetic trajectories---alone or combined with real-world data---substantially outperform all baselines; the strongest baseline, ViNT-Tuned, reaches $\tau=0.67$ at the 20-step HM3D horizon, compared with $0.82$ for VLD. In contrast, the model trained only on real-world videos collapses to a near-constant output, effectively predicting the midpoint of the temporal horizon and yielding $\tau$ near zero. We attribute this to the unstructured nature of internet-scale walking videos, whose motion is often not directed toward a goal; structured, goal-directed simulator trajectories provide the key supervisory signal. On real-world data, the model trained on all sources achieves the best results by a large margin (e.g., $\tau=0.69$ versus $0.44$ for the synthetic-only model on in-the-wild videos at 50 steps) while retaining its Habitat performance. The synthetic-only model consistently ranks second, showing that structured simulator trajectories generalize even to real-world outdoor footage, and that adding diverse real-world video further scales perception to real imagery.

\textbf{Response to goal cues.}
Figure~\ref{fig:goal_cues} evaluates the predictor on recorded frames with a fixed final-frame goal, rather than policy rollouts. Goal cues appear around frame~50 in Habitat and frame~60 in real-world video, where confidence rises and predicted distance falls. VLD exceeds both plotted baselines in trajectory-level $\tau$.

\textbf{Long-horizon consistency.}
To probe settings in which the current observation and the goal share little or no visual overlap, we compute Kendall's~$\tau$ on segments that lie 50--20 and 100--20 steps away from the goal, excluding near-goal frames. VLD retains positive ordinal consistency on HM3D ($\tau=0.40$ and $0.35$, compared with $0.70$ and $0.61$ on the full 50- and 100-step horizons), on in-the-wild videos ($0.42$ and $0.45$), and on embodiment data ($0.52$ and $0.46$), so its ordering persists far from the goal, where direct visual overlap is limited. A pairwise \emph{distance accuracy} test on viewpoints selected to have no visual overlap with the goal, ablations of multi-view inputs and training objectives, and qualitative examples are provided in Appendices~\ref{appdx:vld_ablations}--\ref{appdx:distance_accuracy}.

\FloatBarrier

\subsection{Zero-Shot Substitution in Simulation}
\label{sec:rl_application}

\textbf{Setup.}
We evaluate image-goal navigation in Habitat~\cite{habitat19iccv} on HM3D~\cite{hm3d} and Gibson~\cite{Gibson}. Navigation policies are trained on the standard training splits of HM3D and Gibson and evaluated on their standard validation splits with the standard Habitat evaluation scripts, reporting success rate (SR) and success weighted by path length (SPL). Each policy is evaluated with its training signal (privileged input) and after replacing that signal with a learned distance predictor, keeping the policy weights fixed. As an image-goal baseline, we evaluate FGPrompt-EF~\cite{fgprompt2023}, an end-to-end RL method with leading performance on the Gibson ImageNav benchmark, using its released pretrained checkpoint.

\begin{table}[t]
\centering
\small
\caption{\textbf{Navigation in simulation.} \emph{Left:} each policy evaluated with the privileged distance signal it is trained with. \emph{Right:} image-goal navigation, where the distance input of the same policy is replaced zero-shot by the listed learned predictor with fixed policy weights; the policy in the last distance-based row is trained directly on VLD, and FGPrompt-EF processes the goal image end-to-end. ``+\,conf.'' denotes policies that also receive a confidence input. SR in \%, SPL in $[0,1]$.}
\label{tab:vld_policy_transfer}
\setlength{\tabcolsep}{0pt}
\begin{tabular*}{\linewidth}{@{\extracolsep{\fill}}llcccccccc@{}}
\toprule
& & \multicolumn{4}{c}{\textbf{Privileged input}} & \multicolumn{4}{c}{\textbf{Learned predictor}} \\
\cmidrule(lr){3-6}\cmidrule(l){7-10}
& & \multicolumn{2}{c}{HM3D} & \multicolumn{2}{c}{Gibson} & \multicolumn{2}{c}{HM3D} & \multicolumn{2}{c}{Gibson} \\
\cmidrule(lr){3-4}\cmidrule(lr){5-6}\cmidrule(lr){7-8}\cmidrule(l){9-10}
\textbf{Training signal} & \textbf{Predictor} & SR & SPL & SR & SPL & SR & SPL & SR & SPL \\
\midrule
GT distance (no noise) & -- & 94.7 & 0.582 & 95.8 & 0.610 & -- & -- & -- & -- \\
\textbf{GeoNoise (ours)} & \textbf{VLD} & 90.2 & 0.512 & 92.2 & 0.597 & \textbf{58.7} & \textbf{0.283} & \textbf{76.2} & 0.422 \\
GeoNoise + conf. & VLD & 88.9 & 0.499 & 89.9 & 0.581 & 52.1 & 0.225 & 68.2 & 0.386 \\
OU noise & VLD & 94.4 & 0.543 & 94.5 & 0.646 & 4.9 & 0.011 & 11.5 & 0.040 \\
OU noise + conf. & VLD & 95.6 & 0.601 & 97.3 & 0.721 & 16.5 & 0.078 & 47.1 & 0.315 \\
\midrule
GeoNoise & ViNT-Tuned & -- & -- & -- & -- & 37.6 & 0.143 & 60.5 & 0.256 \\
GeoNoise & VIP-Nav & -- & -- & -- & -- & 15.1 & 0.038 & 27.9 & 0.115 \\
VLD (trained directly) & VLD & -- & -- & -- & -- & 30.8 & 0.186 & 56.6 & 0.423 \\
\midrule
\multicolumn{2}{@{}l}{FGPrompt-EF~\cite{fgprompt2023} (end-to-end)} & -- & -- & -- & -- & 71.6 & 0.393 & 91.1 & 0.588 \\
\bottomrule
\end{tabular*}
\end{table}

\textbf{Zero-shot substitution.}
The GeoNoise-trained policy navigates with VLD predictions without any retraining, reaching $58.7\%$ SR on HM3D and $76.2\%$ on Gibson (Table~\ref{tab:vld_policy_transfer}). Since VLD is never trained on Gibson scenes, the Gibson result also reflects the transfer of the perception module to an unseen environment collection. The same policy reaches $90.2\%$ and $92.2\%$ with its privileged GeoNoise input; with the controller unchanged, this gap isolates the cost of learned perception.

\begin{figure}[!htbp]
\centering
\includegraphics[trim=1160bp 0bp 0bp 815bp,clip,width=0.84\linewidth]{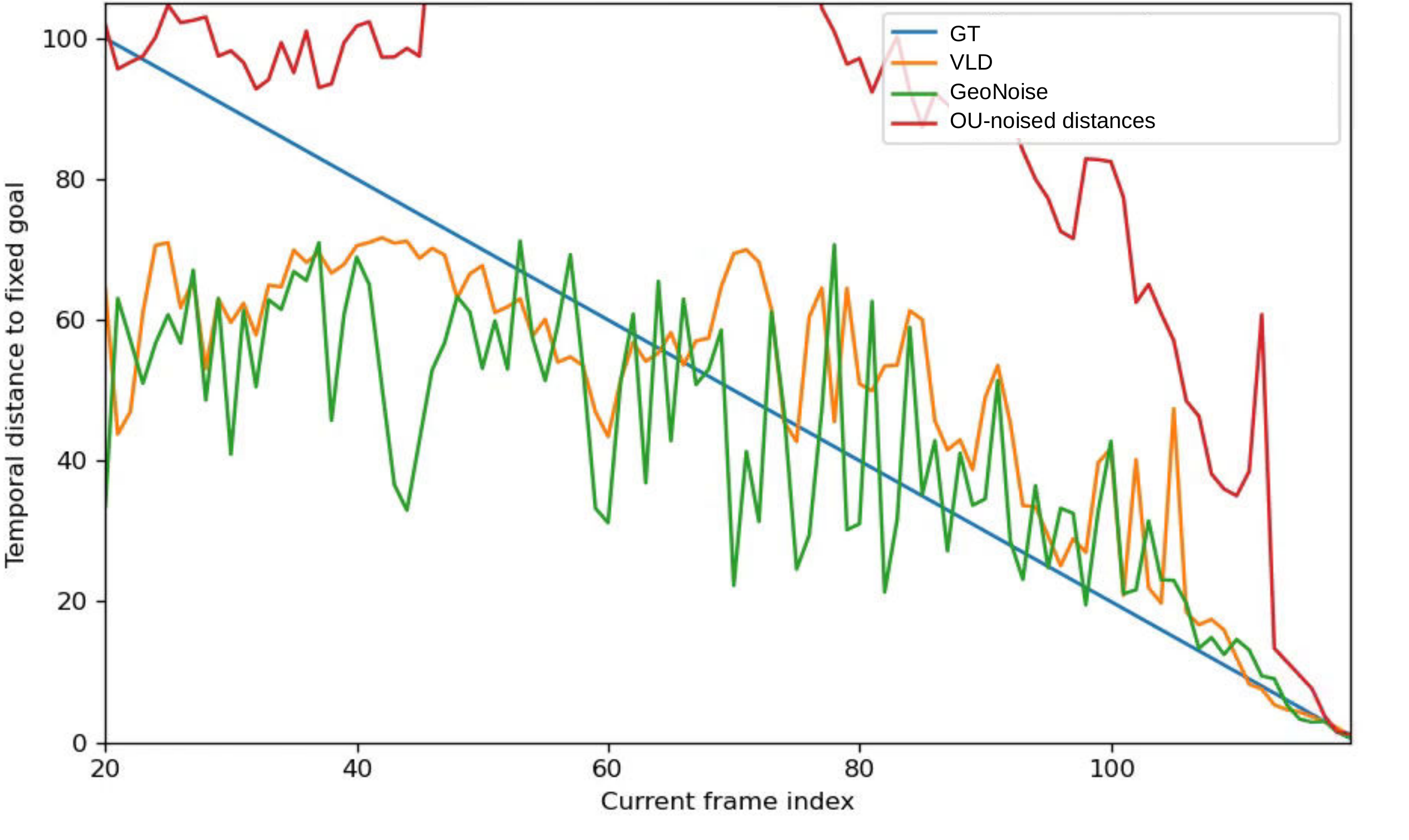}
\caption{\textbf{Noise signals on one trajectory.} GeoNoise (green) follows VLD's (orange) broad scale and trend; OU-noised distances (red) differ. GT (blue) is temporal ground truth; additional examples are in Appendix~\ref{appdx:ou_noise_abl}.}
\label{fig:noise_model_example}
\end{figure}

\textbf{Noise structure determines transferability.}
Policies trained with Ornstein--Uhlenbeck (OU) noise, a generic temporally correlated perturbation that is independent of the scene (Appendix~\ref{appdx:ou_noise_abl}), achieve higher success than GeoNoise-trained policies with their own training signal ($94.4\%$ versus $90.2\%$ on HM3D and $94.5\%$ versus $92.2\%$ on Gibson), yet collapse when VLD is substituted ($4.9\%$ and $11.5\%$). Under the same substitution, GeoNoise-trained policies retain $58.7\%$ and $76.2\%$. Performance under a noisy privileged signal is therefore not a reliable indicator of compatibility with a learned predictor; what matters is that the training signal reproduces the \emph{structure} of the predictor's errors. The OU-perturbed distance does not depend on where the agent looks, whereas GeoNoise, like VLD, depends on viewpoint and visual overlap (Figure~\ref{fig:noise_model_example}); GeoNoise-trained policies learn to use this dependence, scanning and comparing predicted distances across headings before committing to a direction (Appendix~\ref{sec:qual_nav}).

\textbf{Better distance functions yield better navigation.}
Substituting other distance predictors into the same GeoNoise-trained policy follows the ordinal consistency ranking of Table~\ref{tab:ordinal_consistency_all}: ViNT-Tuned reaches $37.6\%$ and $60.5\%$ SR on HM3D and Gibson, and VIP-Nav $15.1\%$ and $27.9\%$, both below VLD. Although GeoNoise is calibrated to VLD, the policy operates with other predictors, and its success follows their ordinal consistency: the interface is modular, and perception can be improved and scaled independently of the controller.

\textbf{Training directly on VLD and confidence inputs.}
Training the policy directly on VLD predictions reaches only $30.8\%$ and $56.6\%$ SR while requiring VLD inference at every simulation step, which makes training substantially more expensive. With GeoNoise training, a confidence input lowers success after substitution ($52.1\%$ and $68.2\%$), so our default policy uses the distance alone; confidence remains part of VLD training (Equation~\eqref{eq:loss}).

\textbf{Comparison with end-to-end image-goal RL.}
FGPrompt-EF, which learns goal-conditioned visual features end-to-end, achieves higher success in simulation ($71.6\%$ on HM3D and $91.1\%$ on Gibson), whereas our policy receives the goal only through a single scalar; most remaining failures of our policy are time-outs during long-horizon exploration (Appendix~\ref{sec:qual_fail_nav}). On about $100$ image-goal episodes in ReplicaCAD~\cite{szot2021habitat2}, on which neither method is trained, the gap shrinks to two points (VLD $70\%$, FGPrompt-EF $72\%$ SR), and on a physical robot the ranking reverses.

\subsection{Real-World Deployment}
\label{sec:real_world}

\textbf{Setup.}
We deploy the GeoNoise-trained policy with VLD, as well as FGPrompt-EF, on a TurtleBot4-Light~\cite{turtlebot4_manual} with an RGB camera mounted approximately 1\,m above the ground, using the same discrete action set as in simulation ($0.5$\,m forward steps) and no real-world fine-tuning. We define five goals in an indoor environment, each specified by an image captured at the goal location (Appendix~\ref{sec:real-wolrd-expr}). The first task starts from a fixed position $S$, and each subsequent task starts at the previous goal; in every task the goal is initially out of view, so the robot must explore before goal cues appear, and two tasks require traversing multiple rooms. We run five trials per goal and method, for a total of 25 trials per method.

\begin{table}[!hbp]
\centering
\small
\setlength{\tabcolsep}{5pt}
\caption{\textbf{Real-world navigation} on a TurtleBot4-Light: successful trials out of five per task (25 per method). Tasks are executed in sequence from the start position $S$. Action counts are reported in Appendix~\ref{sec:real-wolrd-expr}.}
\label{tab:real_world_main}
\begin{tabular}{@{}lcccccc@{}}
\toprule
\textbf{Method} & $S{\to}G_1$ & $G_1{\to}G_2$ & $G_2{\to}G_3$ & $G_3{\to}G_4$ & $G_4{\to}G_5$ & \textbf{Overall} \\
\midrule
VLD + GeoNoise (ours) & \textbf{5/5} & \textbf{4/5} & \textbf{4/5} & 4/5 & \textbf{5/5} & \textbf{22/25 (88\%)} \\
FGPrompt-EF~\cite{fgprompt2023} & 0/5 & 2/5 & 3/5 & \textbf{5/5} & \textbf{5/5} & 15/25 (60\%) \\
\bottomrule
\end{tabular}
\end{table}

\begin{figure*}[!hbp]
\centering
\includegraphics[width=0.9\linewidth]{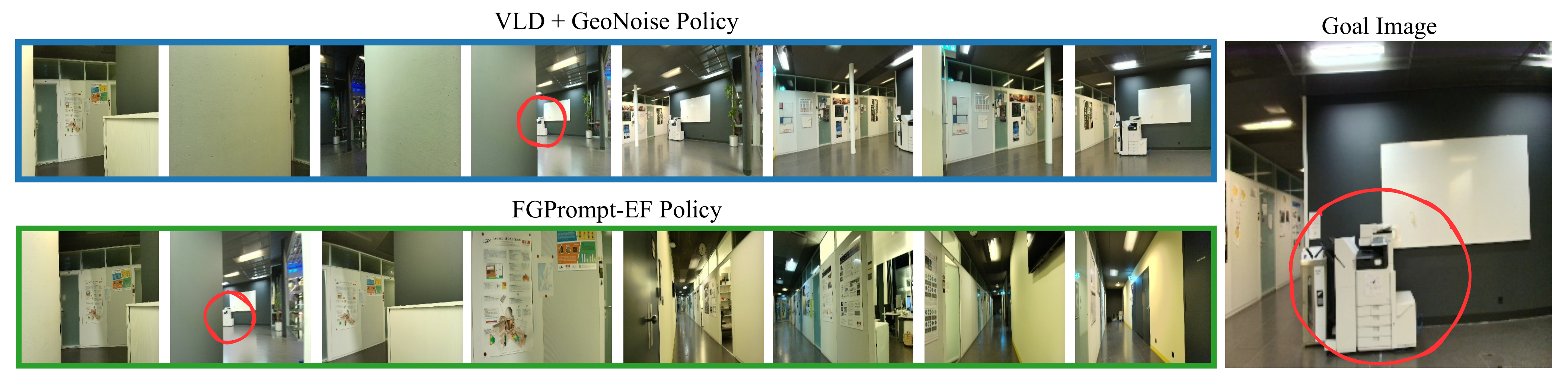}
\caption{\textbf{Real-world trajectory comparison.} Both policies observe the target object (a printer) in one of their views. The VLD-driven policy (top, blue) immediately moves toward the goal as the predicted distance drops sharply, while FGPrompt-EF (bottom, green) initially moves in the opposite direction.}
\label{fig:real_wolrd}
\end{figure*}

\textbf{Results.}
Table~\ref{tab:real_world_main} summarizes the results. The VLD-driven policy succeeds in $22/25$ trials ($88\%$), compared with $15/25$ ($60\%$) for FGPrompt-EF, reversing the ranking observed in simulation. VLD + GeoNoise reaches every goal in at least four of five trials and matches or exceeds FGPrompt-EF on four of the five tasks, whereas the success of FGPrompt-EF varies strongly across tasks, from $0/5$ for $S{\to}G_1$ to $5/5$ for the last two. As in simulation, the VLD-driven policy periodically performs $360^\circ$ scans to compare predicted distances across headings; these rotations cost actions but help the agent pick up weak goal cues and recover from wrong initial choices. FGPrompt-EF, in contrast, tends to commit to a direction early and explores alternatives less often; it frequently stops far from the goal when no goal cues are visible, in particular on $S{\to}G_1$, and occasionally ignores clear goal cues (Figure~\ref{fig:real_wolrd}).

These results support the hypothesis that decoupling perception and control can improve sim-to-real robustness. VLD, trained partly on real-world video, replaces GeoNoise zero-shot as simulation-learned control transfers to the robot. FGPrompt-EF's goal-conditioned features, learned end-to-end in simulation, appear more sensitive to the visual domain shift.

\section{Conclusion}
\label{sec:conclusion}

VLD connects perception learned from video to RL navigation through a scalar goal distance. GeoNoise models viewpoint- and overlap-dependent errors so that VLD replaces the training distance without retraining, whereas generic noise transfers poorly. The physical robot succeeds in $88\%$ of trials versus $60\%$ for an end-to-end baseline that outperforms VLD in simulation. Perception and control can scale independently with more video and simulation.

\clearpage
\bibliography{bibolography}
\clearpage
\appendix
\setcounter{section}{0}
\renewcommand\thesection{\Alph{section}}

\section*{Appendix}

\noindent\textbf{Overview.}
The appendix provides implementation details of VLD and the navigation policy, additional distance-function evaluations, navigation analyses, and the real-world protocol.
\begin{itemize}
  \item \hyperref[appdx:train_algo]{Appendix~\ref*{appdx:train_algo}: distance function training}.
  \item \hyperref[appdx:data_collection]{Appendix~\ref*{appdx:data_collection}: dataset collection} and \hyperref[appdx:yt_links]{Appendix~\ref*{appdx:yt_links}: walking video sources}.
  \item \hyperref[sec:psr_appendix]{Appendix~\ref*{sec:psr_appendix}: projection success ratio} and \hyperref[appdx:reward]{Appendix~\ref*{appdx:reward}: reward function}.
  \item \hyperref[appdx:arch]{Appendix~\ref*{appdx:arch}: alternative architectures} and \hyperref[appdx:vld_setup]{Appendix~\ref*{appdx:vld_setup}: VLD architecture and training setup}.
  \item \hyperref[appdx:vld_ablations]{Appendix~\ref*{appdx:vld_ablations}: distance function ablations} and \hyperref[appdx:losses]{Appendix~\ref*{appdx:losses}: training objectives}.
  \item \hyperref[appdx:qualitative]{Appendix~\ref*{appdx:qualitative}: qualitative distance predictions} and \hyperref[appdx:distance_accuracy]{Appendix~\ref*{appdx:distance_accuracy}: distance accuracy}.
  \item \hyperref[appdx:ou_noise_abl]{Appendix~\ref*{appdx:ou_noise_abl}: distance noise models (GeoNoise and the OU baseline)}.
  \item \hyperref[sec:qual_nav]{Appendix~\ref*{sec:qual_nav}: qualitative navigation}, including \hyperref[sec:qual_succ_nav]{successful rollouts} and \hyperref[sec:qual_fail_nav]{failure cases}.
  \item \hyperref[sec:real-wolrd-expr]{Appendix~\ref*{sec:real-wolrd-expr}: real-world experiments}, with five trials per goal, per-goal results, and physical-robot trajectories.
\end{itemize}

\section{Distance Function Training}
\label{appdx:train_algo}
The training, summarized in Algorithm~\ref{alg:ving}, is framed as a self-supervised regression problem with both positive and negative triplets. The positive examples $(o_i, o_j, td=j-i)$ are formed by sampling $o_i$ and $o_j$ along the same trajectory such that $i \leq j$ and $td = j-i$ is their temporal separation along the recorded trajectory, where $td \leq td_{\max}$. For hard negatives, following ViNG’s negative mining strategy~\cite{VING}, we pair observations from different trajectories and assign the maximum distance $td_{\max}$.
This exposes the model to cross-trajectory and cross-scene mismatches and encourages large predictions for unrelated pairs.

\begin{figure}[b!]
\begin{center}
\includegraphics[width=0.9\linewidth]{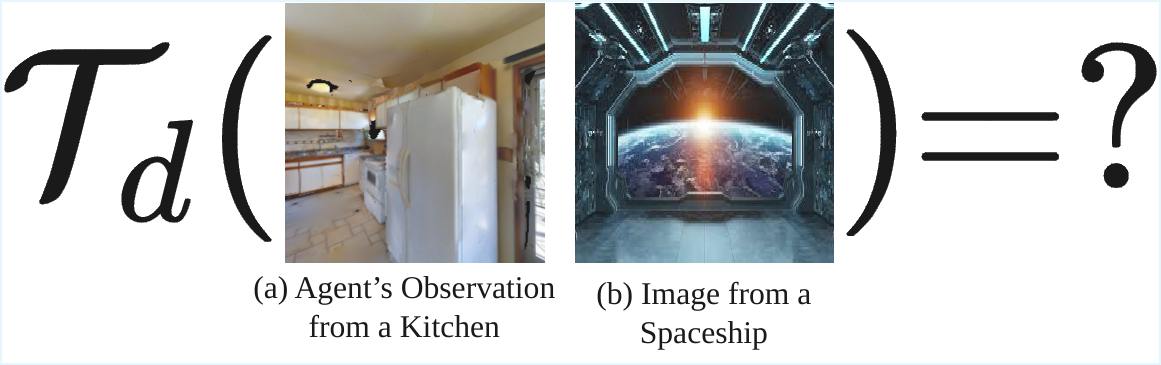}
\end{center}
\caption{\textbf{Negative mining teaches the model to separate unrelated scenes.}
Given two observations from completely different environments (left: kitchen; right: spaceship), the desired behavior is to output a distance close to $td_{\max}$, indicating that the images cannot correspond to nearby states on any trajectory.
Only the model trained with negative mining learns this behavior, predicting a high distance with high confidence; the model trained without it collapses toward an uncertain mid-range prediction.}
\label{fig:negative_mining}
\end{figure}

We further highlight the role of \emph{negative mining} through the example shown in Figure~\ref{fig:negative_mining}. When computing the distance between two observations—one taken in a standard indoor kitchen and the other onboard a spaceship—we expect the model to assign a value close to $td_{\max}$, reflecting that these views are entirely unrelated. In practice, comparing models trained with and without negative mining shows a sharp contrast. The model trained with negative mining predicts a distance of approximately 102, slightly exceeding its $td_{\max}=100$, consistent with the maximum-distance target assigned to unrelated pairs. In contrast, the model trained without negative mining predicts a distance of only 50, essentially the midpoint of the range. Moreover, the negative-mining-trained model produces a confidence score close to 1.0, while the model without negative mining yields only 0.13, illustrating how negative mining teaches the model to confidently separate unrelated scenes.

\begin{algorithm}
  \caption{Distance function training}
  \label{alg:ving}
  \begin{algorithmic}[1]
    \Require Trajectories $\{\tau^{(k)}=(o^{(k)}_1,o^{(k)}_2,\dots)\}_{k=1}^{K}$,
    max horizon $td_{\max}$, negative sampling probability $p_{\text{neg}}$, parametrized distance model $\mathcal{T}_\theta(o, g)$, loss function $\mathcal{L}$
    \Statex

    \State \textbf{Construct positives:}
    $\mathcal{D}_+ \gets \{(o^{(k)}_i,\,o^{(k)}_j,\, td=j-i) \;\mid\; k\in\{1,\ldots,K\},\; 1\le i\le j,\; j-i\le td_{\max}\}$

    \State \textbf{Construct negatives:}
    $\mathcal{D}_- \gets
    \{(o^{(k)}_i,\,o^{(\ell)}_j,\, td=td_{\max}) \;\mid\; k\ne \ell\}$

    \State Initialize distance model $\mathcal{T}_\theta(o, g) \to (\hat{t}, \hat{c})$

    \Statex

    \While{not converged}
      \If{$\text{Uniform}(0, 1) < p_{\text{neg}}$} \Comment{negative step}
        \State Sample mini-batch $\mathcal{B} \sim \mathcal{D}_-$
      \Else \Comment{positive step}
        \State Sample mini-batch $\mathcal{B} \sim \mathcal{D}_+$
      \EndIf

      \State Compute predictions $\{(\hat{t}_n, \hat{c}_n)\} \gets \{\mathcal{T}_\theta(o_n, g_n)\;|\;(o_n,g_n,td_n)\in\mathcal{B}\}$
      \State Compute loss: $\mathcal{L}(\theta) \gets \text{LossFunction}(\{(\hat{t}_n,\hat{c}_n,td_n)\}_{n\in\mathcal{B}})$
      \State Update parameters: $\theta \leftarrow \theta - \eta \nabla_\theta \mathcal{L}(\theta)$
    \EndWhile

    \State \Return trained distance function $\mathcal{T}_\theta$
  \end{algorithmic}
\end{algorithm}

\section{Dataset Collection}
\label{appdx:data_collection}

Training data for the distance function is collected from three kinds of sources: (i) synthetic trajectories from simulators, (ii) “in-the-wild” internet-scale videos, and (iii) embodiment datasets collected on real robots. A summary is provided in Table~\ref{tab:datasets}.

Since these datasets span different embodiments and capture rates, the average velocity of agents varies considerably. To normalize the effective temporal spacing, we adjust the frame rate so that the distance covered between two consecutive frames is approximately $0.25$\,m. This makes the supervision units comparable across sources. When velocity metadata is unavailable, we estimate it using a simple visual odometry method~\cite{vo}.

The collection procedure for each data type:

\begin{enumerate}
    \item \textbf{Synthetic data.}
    We use Habitat~\cite{habitat19iccv} with the HM3D scenes~\cite{hm3d}, which provide well-structured trajectories together with specific goal object annotations and corresponding image goals~\cite{instanceobjectgoal}. We exploit this structure by generating trajectories with the \emph{ShortestPathFollower}. To make goal images more meaningful, we extend the follower: instead of stopping at the exact point-goal location, the agent continues until the goal object is visible and then rotates to orient toward it before issuing the stop action. This ensures that the final goal view clearly depicts the object of interest.

    This procedure yields 475,562 training trajectories from the 145 scenes of the standard HM3D training split and 2,271 validation trajectories from 35 HM3D validation scenes after filtering very short routes. We additionally generate 994 validation trajectories in the Gibson validation scenes, which are never used for VLD training. Navigation policies are trained and evaluated separately on the standard image-goal navigation episodes (Section~\ref{sec:rl_application}).

    \item \textbf{``In-the-wild'' videos.}
    Following CityWalker~\cite{city}, we curate 1,720 hours of YouTube videos of humans walking across different outdoor settings, seasons, and times of day. Sources of these videos are listed in Section~\ref{appdx:yt_links}. We also add 50 hours of diverse sequences from Ego4D~\cite{ego4d}, filtered to scenarios such as ``street walking'', ``indoor navigation'', and ``jogging/cycling''. No further preprocessing is applied beyond adjusting the frame rate. For both sources, we hold out roughly 10\% of trajectories for validation.

    \item \textbf{Embodiment datasets.}
    We include several smaller but very high-quality datasets collected on physical robots. SCAND~\cite{scanD} provides trajectories from Spot and Jackal across indoor and outdoor settings. The  GrandTour~\cite{GTDataset} spans more than 49 environments, including indoor, urban, natural, and mixed settings collected with a quadrupedal robot. We also incorporate CityWalker’s own robot-collected subset~\cite{city}. These datasets are limited in scale but highly diverse, and we upsample them during training. Again, around 10\% of trajectories are reserved for validation.
\end{enumerate}

\begin{table}[t]
\caption{Datasets used for training and validating the temporal distance function. We report average velocity, total hours, and dataset type.}
\label{tab:datasets}
\begin{center}
\small
\begin{tabular}{lccc}
\multicolumn{1}{c}{\bf Dataset} &
\multicolumn{1}{c}{\bf Avg. Velocity} &
\multicolumn{1}{c}{\bf Total Hrs} &
\multicolumn{1}{c}{\bf Type}
\\ \hline \\[-0.9em]
HM3D        & $2.5$ m/s       & 1150h & Synthetic     \\
YouTube     & $0.5$ m/s       & 1720h & In-the-wild   \\
Ego4D       & $0.5$--$1.5$ m/s& 50h   & In-the-wild   \\
SCAND       & $1.5$--$2$ m/s  & 9h    & Embodiment    \\
Grand Tour  & $1$ m/s         & 5h    & Embodiment    \\
CityWalker  & $1.5$ m/s       & 15h   & Embodiment    \\
\hline \\[-0.9em]
\multicolumn{2}{r}{\bf Total} & 2949h & \\
\end{tabular}
\end{center}
\end{table}

\section{Walking Data Video Sources}
\label{appdx:yt_links}

For the in-the-wild data component, we follow the CityWalker dataset~\cite{city} and use the same YouTube walking video sources. The playlists are publicly available and can be accessed at:

\begin{itemize}
  \item \href{https://www.youtube.com/playlist?list=PLsRHC_C5DDaP3aur0aHp1FsXr3tMY0sDh}{Day Walking Tours}
  \item \href{https://www.youtube.com/playlist?list=PLsRHC_C5DDaM21SwCw7UJx1W_1uwCt3yq}{Sunset Walking Tours}
  \item \href{https://www.youtube.com/playlist?list=PLsRHC_C5DDaPJiB6bLTxmWjRK0A1eOB5r}{Rainy Walking Tours}
  \item \href{https://www.youtube.com/playlist?list=PLsRHC_C5DDaOw4ByXDOULuqdT08txXKHQ}{Night City Walking Tours}
  \item \href{https://www.youtube.com/playlist?list=PLsRHC_C5DDaOhAoKRbmf26MEDqU5EqRo_}{Snow Days}
\end{itemize}

These playlists comprise almost 2000 hours of human walking footage captured across diverse urban and natural environments, recorded at various times of day and under varying weather conditions. They serve as the source for our internet-scale “in-the-wild” trajectories.

\section{Projection Success Ratio (PSR)}
\label{sec:psr_appendix}

The projection success ratio (PSR) provides a geometric proxy for visual overlap between a goal view and the agent’s current egocentric view. It is defined as the fraction of valid goal pixels that project consistently into the current depth image:
\begin{equation}
\mathrm{PSR} = \frac{1}{N}\sum_{p=1}^N \mathbf{1}\!\left(\; \big|Z^{\text{curr-pred}}_p - Z^{\text{curr}}_{\pi(p)}\big| < \tau \;\right),
\end{equation}
where $N$ is the number of valid depth pixels in the goal image, $Z^{\text{curr-pred}}_p$ is the predicted depth of pixel $p$ after unprojection from the goal frame and transformation into the current camera frame, $\pi(p)$ is its projection into the current image plane using calibrated intrinsics, and $Z^{\text{curr}}_{\pi(p)}$ is the observed depth at the corresponding pixel in the current view. The threshold $\tau$ controls the tolerance for depth consistency.

\begin{figure*}[t]
\centering

\begin{subfigure}{0.9\linewidth}
  \centering
  \includegraphics[width=\linewidth]{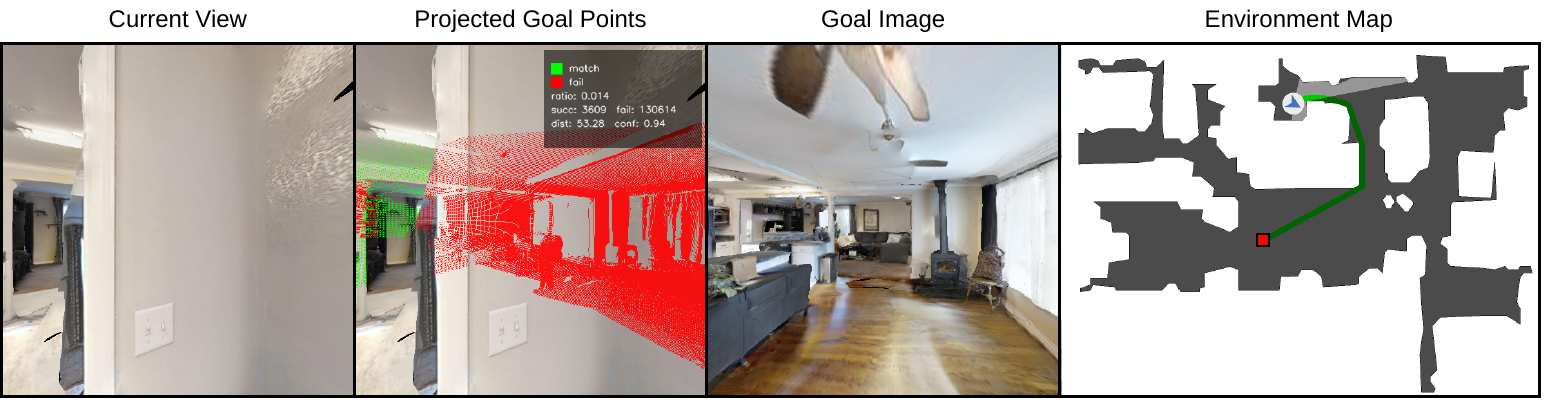}
  \caption{\textbf{No overlap.} The goal view is occluded (e.g., by a wall), yielding a near-zero projection success ratio.}
\end{subfigure}

\vspace{0.5em}

\begin{subfigure}{0.9\linewidth}
  \centering
  \includegraphics[width=\linewidth]{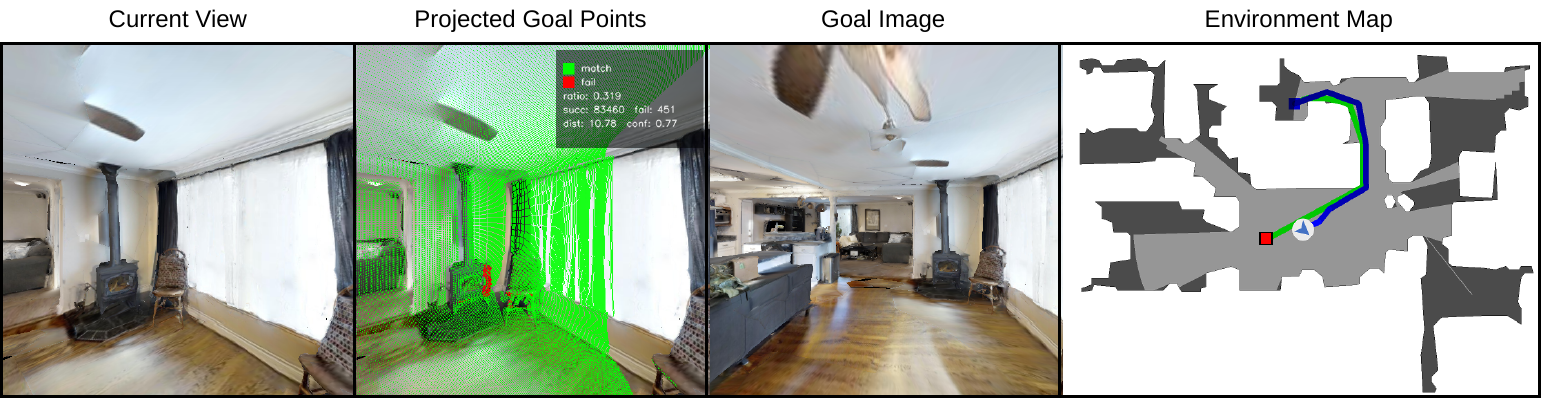}
  \caption{\textbf{Partial overlap.} A subset of projected points land consistently, producing an intermediate projection success ratio.}
\end{subfigure}

\vspace{0.5em}

\begin{subfigure}{0.9\linewidth}
  \centering
  \includegraphics[width=\linewidth]{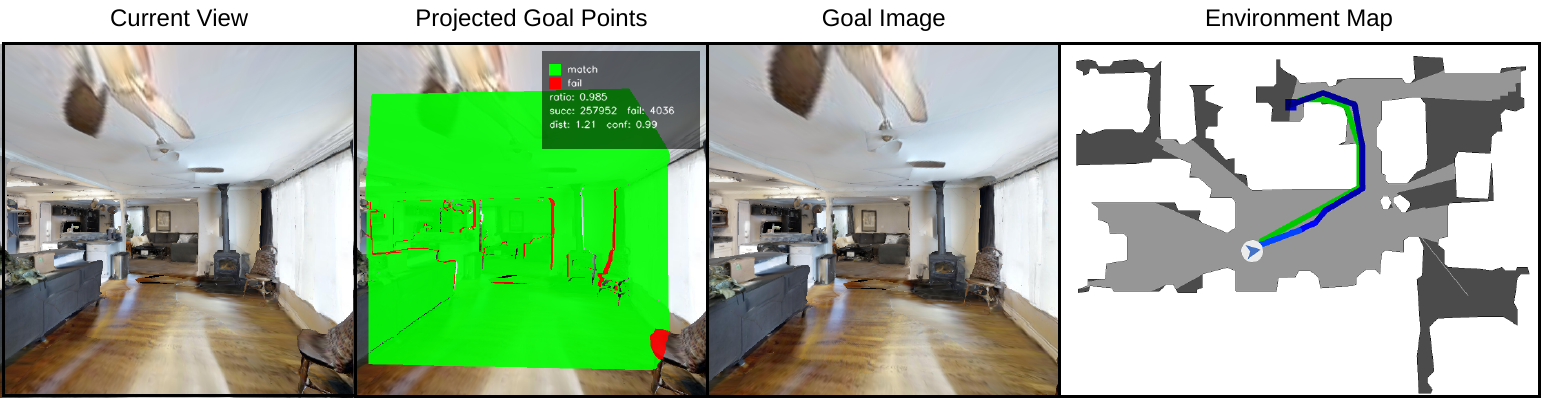}
  \caption{\textbf{Full overlap.} Most projected points are consistent, resulting in a high projection success ratio.}
\end{subfigure}

\caption{\textbf{Projection success ratio (PSR) visualization.}
Each row shows the current view, the goal-image points projected into the current view (green: depth-consistent matches, red: failed matches), the goal image, and a top-down map of the environment.
The legend in the second panel reports the PSR $r$, the numbers of matched and failed points, and the predicted distance $\hat d$ and confidence $\hat c$ of GeoNoise.}
\label{fig:psr_viz}
\end{figure*}

Figure~\ref{fig:psr_viz} illustrates PSR in practice. With no overlap, nearly all projected pixels fail to match (yielding PSR $\approx 0$). With partial overlap, a subset of projections is consistent, resulting in an intermediate PSR. With full overlap, most goal pixels successfully project, producing a high PSR $\approx 1$.

\section{Reward Function}
\label{appdx:reward}

For policy training, we extend the \emph{Zero-Exploration Reward (ZER)} formulation~\cite{zer} to explicitly encourage the agent to orient toward the goal. At each step $t$, the agent’s overlap with the goal is quantified by the projection success ratio $psr_t \in [0,1]$ (Appendix~\ref{sec:psr_appendix}). Let $psr_t^{\max}$ denote the maximum overlap achievable at the current position if the agent were rotated optimally, and let $psr_s$ be a success threshold. The reward is then given as
\begin{equation}
r_t = (d_{t-1} - d_t) - \gamma\,(1 - psr_t) + \mathbbm{1}\bigl[\,d_t < d_s \land psr^{\max}_t > psr_s\,\bigr]\,R_s,
\label{eq:reward_ratio}
\end{equation}
where $d_t$ is the true geometric distance-to-goal at time $t$, $\gamma$ is a step penalty, $d_s$ is a success margin, and $R_s$ is a success bonus applied when both the distance and overlap conditions in the indicator are satisfied.

\section{Alternative Distance Function Architectures}
\label{appdx:arch}

\begin{figure}[t]
\begin{center}
\includegraphics[width=1.\linewidth]{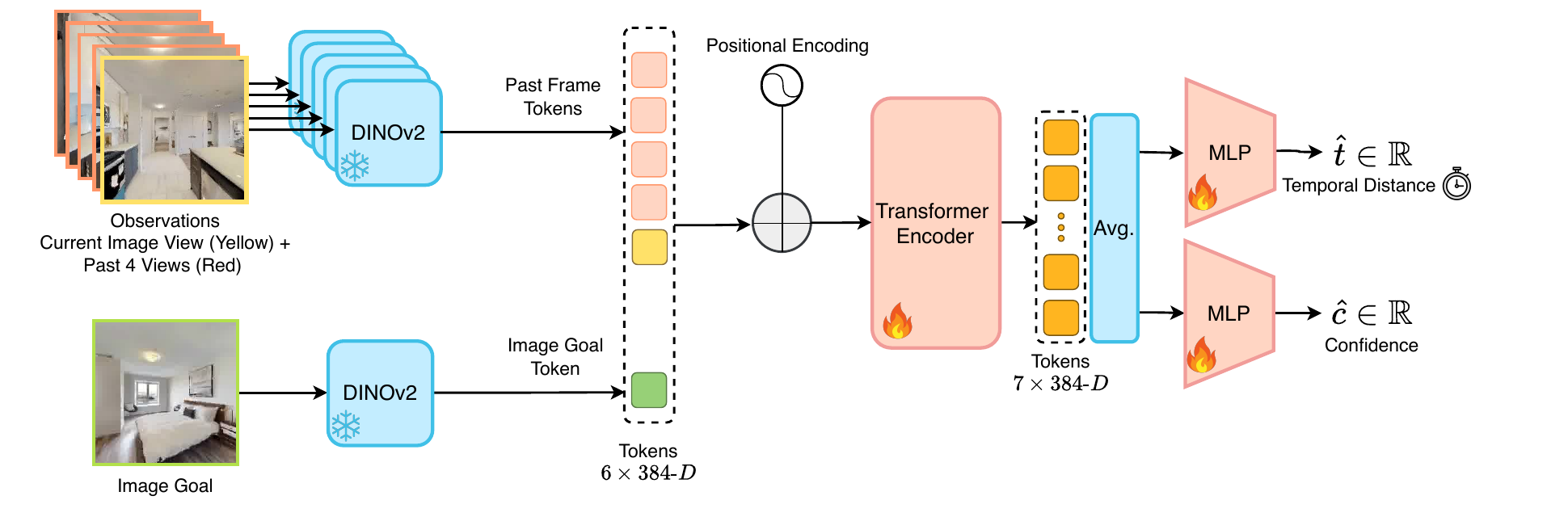}
\end{center}
\caption{\textbf{Encoder-based alternative.} Observation and goal tokens are concatenated and processed jointly with a Transformer encoder, following the ViNT design but using frozen DINOv2 encoders.}
\label{fig:alt-encoder}
\end{figure}

\begin{figure}[t]
\begin{center}
\includegraphics[width=1.0\linewidth]{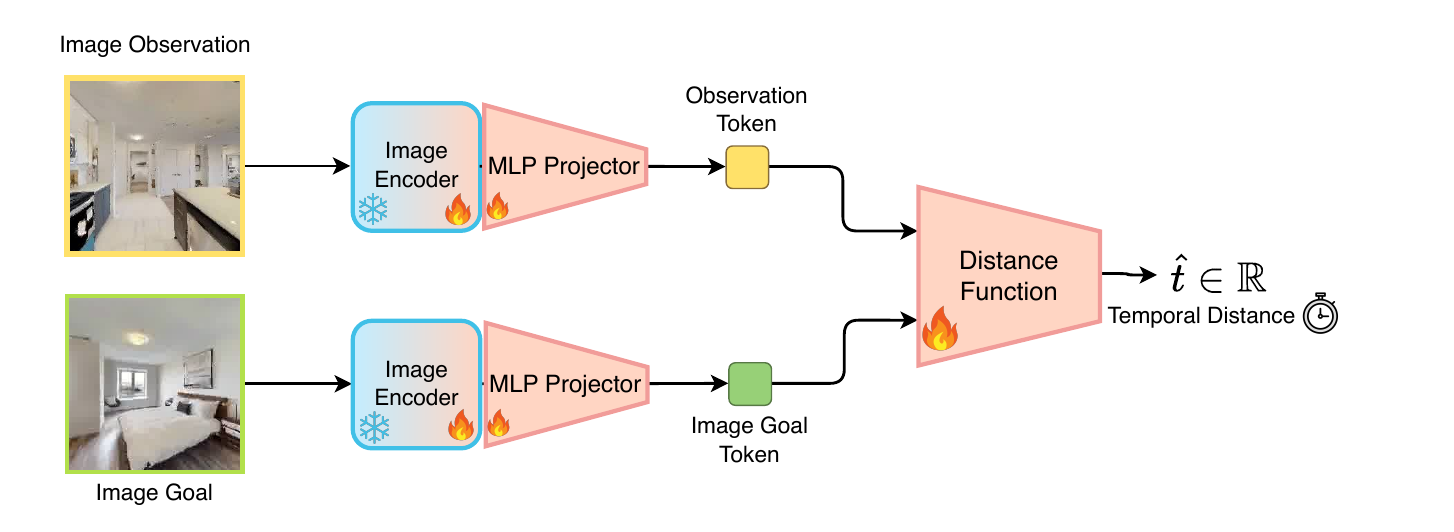}
\end{center}
\caption{\textbf{VIP-style alternatives.} Observations and goals are encoded separately via some image encoder that can either be frozen or trainable, and distance is computed directly in the embedding space (via $\ell_2$, linear, or quasimetric neural networks).}
\label{fig:alt-vip}
\end{figure}

To test the robustness of our design choices, we evaluate several alternative architectures for distance prediction, summarized in Figures~\ref{fig:alt-encoder} and \ref{fig:alt-vip}. These span both encoder-based and embedding-distance approaches.

\begin{figure}[t]
\begin{center}
\includegraphics[width=1.\linewidth]{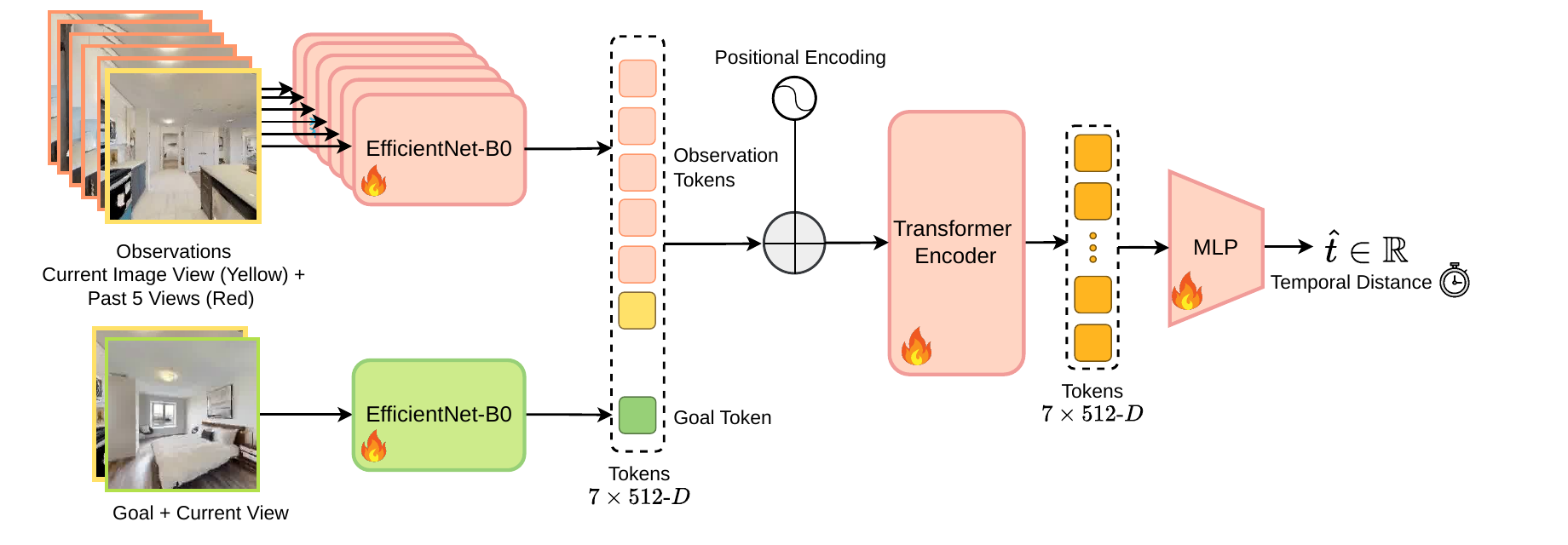}
\end{center}
\caption{\textbf{Adapted ViNT Model Architecture~\cite{vint}.} We modify the original ViNT model by removing the action prediction head and retaining only the temporal distance prediction branch. The model encodes past observations and goal images with two EfficientNet-B0 encoders, fuses the resulting features into a sequence of tokens, and processes them with a Transformer to predict the (temporal) distance to the goal.}
\label{fig:vint_arch}
\end{figure}

\paragraph{(i) ViNT-style baseline.}
We directly adopt the ViNT architecture~\cite{vint}, where observation and goal tokens are concatenated and processed jointly by a Transformer encoder (see Figure~\ref{fig:vint_arch}). In contrast to our use of frozen internet-scale encoders (DINOv2), ViNT relies on EfficientNet backbones~\cite{effNet} that are trained from scratch. For consistency with our setting, we remove the action-prediction head and retain only the temporal distance output. We evaluate both the released ViNT model (pre-trained on their curated datasets) and variants trained on our data. This serves as a natural baseline, since ViNT also predicts temporal distance as an auxiliary signal, though its backbone and training setup differ substantially from ours.

\paragraph{(ii) Encoder over joint tokens.}
As shown in Figure~\ref{fig:alt-encoder}, we experiment with an encoder-only Transformer applied to the concatenation of observation and goal embeddings. Unlike our decoder-based design, which retains the full token sequence from DINOv2, here each image is reduced to a single pooled embedding from its backbone. These embeddings are then concatenated, augmented with positional encodings, and processed symmetrically by a Transformer encoder. The output tokens are then averaged, and the resulting embedding is used to regress distance and confidence. This design allows us to isolate the effect of our decoder-based query–memory formulation versus pooled joint encoding.

\paragraph{(iii) VIP-style embedding distances.}
Inspired by VIP~\cite{vip}, we also test architectures where observation and goal images are encoded independently, and temporal distance is predicted as a function of their embeddings (Figure~\ref{fig:alt-vip}). Concretely, we experiment with simple $\ell_2$ norms as well as learned functions such as a linear layer with ReLU. Since the prediction head in this setup is relatively weak and lacks the capacity of a decoder with cross-attention, the choice of image encoder becomes especially critical. We therefore evaluate both frozen and trainable encoders, ranging from lightweight CNNs to DINOv2. These baselines use the same image-goal setting as VLD.

\paragraph{(iv) Quasimetric neural networks.}
Finally, we consider using the distance head with a quasimetric neural network (QNN)~\cite{iqe}, which enforces quasimetric constraints in the embedding space by design. This enforces a triangle inequality while permitting asymmetric distances, aligning with our QRL-inspired objective (Section~\ref{appdx:losses}). We embed this into the VIP-style architecture of Figure~\ref{fig:alt-vip}.

\section{VLD Architecture and Training Setup}
\label{appdx:vld_setup}
Tables~\ref{tab:hyperparams-model} and~\ref{tab:hyperparams-training} summarize the hyperparameters of our Single-View VLD model and training setup. We set the maximum temporal distance to $td_{\max}=100$ steps, corresponding to approximately $25$ meters in Habitat, and train using the inlier--outlier Gaussian mixture NLL loss, with the reliable and outlier variances fixed at $\sigma_R^2=4$ and $\sigma_O^2=40$, respectively.

The model contains approximately $33$M parameters: a frozen DINOv2-Small backbone (approximately $22$M) and trainable prediction modules (approximately $11$M).

The Multi-View variant differs only in that it includes a temporal pooling module consisting of a single self-attention layer across $T=5$ frames (the current observation and four preceding frames) (Figure~\ref{fig:vld_full}), which adds roughly 2M trainable parameters. Unless otherwise stated, all experiments reported under ``VLD'' refer to the Single-View variant as our default comparison model.

\begin{table}[t]
\caption{Hyperparameters for the Single-View VLD model.}
\label{tab:hyperparams-model}
\centering
\small
\setlength{\tabcolsep}{7pt} 
\begin{tabular}{l r}
\multicolumn{2}{c}{\bf VLD Model} \\
\hline \\[-0.9em]
Total \# Parameters (approx.)  & 33M \\
Trainable \# Parameters        & 11M \\
Image Encoder                  & DINOv2-Small \\
Image Input Res.               & $224 \times 224$ \\
Image Token Dimension          & $384$ \\
Attn. Hidden Dim.              & $384$ \\
\# Decoder Layers              & 4 \\
\# Attention Heads             & 8 \\
Activation Function            & GELU \\
\end{tabular}
\end{table}

\begin{table}[t]
\caption{Training setup for the Single-View VLD model.}
\label{tab:hyperparams-training}
\centering
\small
\setlength{\tabcolsep}{7pt} 
\begin{tabular}{l r}
\multicolumn{2}{c}{\bf Training} \\
\hline \\[-0.9em]
\# Steps                       & 2M \\
Batch Size                     & 64 \\
Learning Rate                  & $5 \times 10^{-5}$ \\
Optimizer                      & AdamW \\
LR Schedule                    & Cosine \\
Warmup Steps                   & 10K \\
Max temporal dist. ($td_{\max}$)   & 100 \\
Neg. mining prob. ($p_{\text{neg}}$) & 0.05 \\
Conf. “reliable” var. ($\sigma_R^2$) & 4 \\
Conf. “outlier” var. ($\sigma_O^2$)  & 40 \\
Compute Resources              & $4 \times$ RTX 3090 \\
Training Walltime              & 120 h \\
\end{tabular}
\end{table}

\section{Distance Function Ablations}
\label{appdx:vld_ablations}

Beyond the experiments with training data presented in Section~\ref{sec:ordinal_eval} in the main paper body, which compared synthetic, real-world, and combined training sources, we perform additional ablations to examine two key design choices: (i) the impact of using multi-view versus single-view inputs, and (ii) the effect of different learning objectives.

\subsection{Multi-view inputs}
\label{appdx:multiview}
\begin{figure*}[t]
\centering
\includegraphics[width=0.95\linewidth]{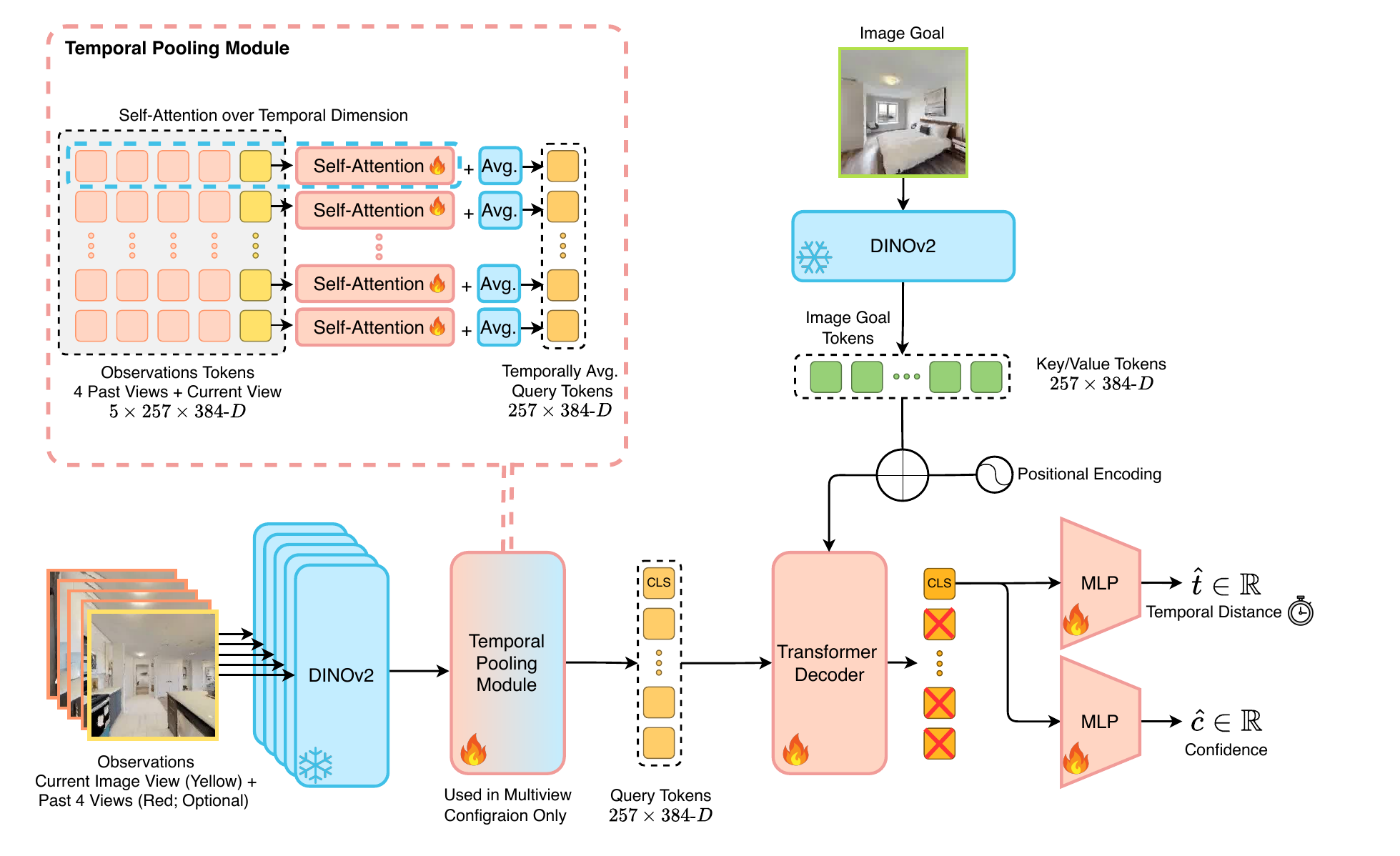}
\caption{\textbf{Multi-view Visual Learned Distance (VLD) architecture.}
Similar to the single-view variant (Figure~\ref{fig:vld}), but processes a short temporal window of past views. Each frame is encoded with DINOv2, followed by self-attention across temporal tokens before aggregation and decoding.}
\label{fig:vld_full}
\end{figure*}

Drawing inspiration from ViNT~\cite{vint}, which leverages multiple past frames for improved spatial reasoning, we extend our VLD architecture to support multi-view inputs, shown in Figure~\ref{fig:vld_full}.
The model independently encodes a sequence of recent observations using DINOv2, applies self-attention across temporal tokens, and averages the resulting representations before decoding them into the final scalar distance prediction.

\textbf{Results on Habitat datasets.}\quad
Table~\ref{tab:ordinal_consistency_habiatat_multiview} compares single-view and multi-view VLD models.
The multi-view variant matches the single-view model at the 20-step HM3D horizon and achieves slightly higher ordinal consistency in the remaining settings.
However, the improvements are relatively small compared to the additional computational cost, suggesting that temporal context offers modest robustness gains while the single-view design remains a more efficient choice for most downstream settings.

\begin{table}[t]
  \centering
  \footnotesize
  \caption{\textbf{Single-view vs.\ multi-view inputs.} Ordinal consistency (Kendall’s~$\tau$, ↑ higher is better) of VLD models on Habitat datasets. We report results at 20, 50, and 100-step horizons.}
  \label{tab:ordinal_consistency_habiatat_multiview}
  \begin{tabular}{@{}lcccccc@{}}
    \toprule
    \multirow{2}{*}{\textbf{Model}} &
    \multicolumn{3}{c}{\textbf{HM3D (↑)}} &
    \multicolumn{3}{c}{\textbf{Gibson (↑)}} \\
    \cmidrule(lr){2-4} \cmidrule(lr){5-7}
    & 20 & 50 & 100 & 20 & 50 & 100 \\
    \midrule
    VLD (single) & \textbf{0.82} & 0.70 & 0.61 & 0.84 & 0.73 & 0.71 \\
    VLD (multi)  & \textbf{0.82} & \textbf{0.71} & \textbf{0.63} & \textbf{0.86} & \textbf{0.75} & \textbf{0.73} \\
    \bottomrule
  \end{tabular}
\end{table}

\subsection{Learning objectives}
\label{appdx:objectives}

We further compare our Gaussian mixture negative log-likelihood (NLL) objective against several alternatives:
(i) mean squared error (MSE),
(ii) the implicit time-contrastive loss used in VIP~\cite{vip}, and
(iii) the QRL loss~\cite{qrl}.
Full formulations are provided in Appendix~\ref{appdx:losses}.

\textbf{Results on Habitat datasets.}\quad
Table~\ref{tab:ordinal_consistency_objective} summarizes the results.
The Gaussian mixture NLL achieves the highest ordinal consistency in five of six settings and ties MSE at the 100-step HM3D horizon, which supports modeling confidence during distance learning.
In contrast, QRL exhibits unstable behavior—tending to saturate and under-predict large distances.

\begin{table}[t]
  \centering
  \footnotesize
  \caption{\textbf{Comparison of learning objectives.} Ordinal consistency (Kendall’s~$\tau$, ↑ higher is better) of VLD models trained with different objectives on Habitat datasets. Results shown for 20, 50, and 100-step horizons.}
  \label{tab:ordinal_consistency_objective}
  \begin{tabular}{@{}l@{\hspace{4pt}}ccc@{\hspace{4pt}}ccc@{}}
    \toprule
    \multirow{2}{*}{\textbf{Objective}} &
    \multicolumn{3}{c}{\textbf{HM3D (↑)}} &
    \multicolumn{3}{c}{\textbf{Gibson (↑)}} \\
    \cmidrule(lr){2-4} \cmidrule(lr){5-7}
    & 20 & 50 & 100 & 20 & 50 & 100 \\
    \midrule
    VLD (NLL) & \textbf{0.82} & \textbf{0.70} & \textbf{0.61} & \textbf{0.84} & \textbf{0.73} & \textbf{0.71} \\
    VLD (MSE) & 0.71 & 0.67 & \textbf{0.61} & 0.78 & 0.70 & 0.68 \\
    VLD (QRL) & 0.17 & 0.20 & 0.16 & 0.06 & 0.07 & 0.07 \\
    VLD (VIP) & 0.73 & 0.53 & 0.43 & 0.79 & 0.59 & 0.56 \\
    \bottomrule
  \end{tabular}
\end{table}

\section{Training objectives}
\label{appdx:losses}

Although our main experiments rely on the Gaussian mixture NLL, we also experiment with alternative objectives that allow for self-supervised training and are conceptually attractive because they link the temporal distance function more directly to (i) reward learning from large-scale video (VIP)~\cite{vip} and (ii) geometric consistency of value functions (QRL)~\cite{qrl}.

As a baseline, following ViNG and ViNT, we first consider the simplest formulation: minimizing mean squared error (MSE) between predicted and ground-truth temporal distances~\cite{VING, vint}. This approach is straightforward and often surprisingly effective, but it does not explicitly encourage the distance function to acquire useful geometric or representational structure. We therefore explore objectives designed specifically for temporal distance learning.

\textbf{Value-Implicit Pretraining (VIP).}
VIP~\cite{vip} was introduced as a way to learn temporal distance functions from large-scale human video by framing representation learning as an \emph{implicit goal-conditioned RL problem}. Rather than regressing absolute distances, VIP trains a visual encoder such that the distance to the goal is reflected in the similarity of embeddings. Concretely, the learning objective is designed such that initial and goal frames are pulled closer together in embedding space, while intermediate frames are repelled through a temporal-difference style objective (see Algorithm~\ref{alg:vip}). This produces embeddings that are smooth over time and encode a notion of goal progress.

The appeal of VIP in our context is that it provides a temporally consistent signal that does not depend on explicit action labels, aligning well with our emphasis on large-scale video pretraining. However, it is worth noting that VIP has been predominantly evaluated in manipulation domains, where image overlap between successive frames is high and the visual signal remains relatively consistent. In contrast, navigation often involves long horizons with little or no visual overlap, so we test whether VIP’s benefits transfer to this more challenging setting.

\begin{algorithm}[t]
  \caption{Value-Implicit Pretraining (VIP)~\cite{vip}}
  \label{alg:vip}
  \begin{algorithmic}[1]
    \Require Offline videos $D = \{(o^i_1,\dots,o^i_{h_i})\}_{i=1}^N$, visual encoder $\phi$
    \For{number of training iterations}
      \State Sample sub-trajectories $\{o^i_t, \dots, o^i_k, o^i_{k+1}, \dots, o^i_T\}$ from $D$
      \State Compute loss:
        \[
        \begin{aligned}
        \mathcal{L}(\phi) = &\; \frac{1-\gamma}{B}\sum_{i=1}^B
        \|\phi(o^i_t) - \phi(o^i_T)\|_2 \\[4pt]
        &+ \log \frac{1}{B} \sum_{i=1}^B \exp\big(
        \|\phi(o^i_k)- \phi(o^i_T)\|_2 \\[4pt]
        &-\gamma \|\phi(o^i_{k+1}) - \phi(o^i_T)\|_2
        \big)
        \end{aligned}
        \]

      \State Update $\phi$ using SGD:  $\phi \gets \phi - \alpha_\phi \nabla \mathcal{L}(\phi)$
    \EndFor
  \end{algorithmic}
\end{algorithm}

\textbf{Quasimetric RL (QRL).}
Inspired by Quasimetric RL (QRL)~\cite{qrl}, we also consider training the distance function to satisfy \emph{quasimetric} properties. A quasimetric is a generalization of a metric that allows asymmetry (i.e., $d(x,y) \neq d(y,x)$) while still respecting the triangle inequality. QRL shows that in multi-goal RL, the optimal value function is always a quasimetric, making these geometric constraints both natural and theoretically well-founded.

Formally, let $d_\theta(o,g)\!\in\![0,D_{\max}]$ denote the predicted temporal distance from observation $o$ to goal $g$. The QRL objective optimizes a Lagrangian that (i) \emph{spreads} random pairs apart up to a margin $D_{\max}$ and (ii) enforces \emph{local one-step consistency} on consecutive frames, without requiring any action labels (self-supervised learning objective):

\begin{align}
\label{eq:qrl-ours}
\min_{\theta}\; \max_{\lambda\ge 0}\;
&\underbrace{\mathbb{E}_{o\sim p_{\text{state}},\,g\sim p_{\text{goal}}}
\big[\;\psi\!\big(D_{\max}-d_\theta(o,g)\big)\;\big]}_{\text{global spreading}} \nonumber \\
&\;+\;
\lambda\!\Bigg(
\underbrace{\mathbb{E}_{(o,o')\sim p_{\text{trans}}}\!\left[
\big(\mathrm{relu}\!\left(d_\theta(o,o')-c\right)\big)^{\!2}\right]}_{\text{local one-step constraint}}
-\varepsilon^{2}\Bigg).
\end{align}

where $\psi(x)=\mathrm{softplus}(x)$, $c$ is the per-step cost (equal to the temporal distance between $o$ and $o'$), and $\varepsilon>0$ is a slack tolerance. In practice, we parameterize $\lambda=\mathrm{softplus}(\rho)$, clamp it to a maximum value (to prevent degenerate solutions where the model becomes overly conservative), and apply a gradient-reversal trick on $\rho$ to implement the $\max_{\lambda\ge 0}$ update.

The \emph{global spreading} term encourages random $(o,g)$ pairs to lie near the maximum margin $D_{\max}$, ensuring broad separation of states. The \emph{local constraint} enforces that the predicted distance between neighboring observations\footnote{$o'$ is sampled such that the distance between $o$ and $o'$ is less than a set margin, $c\leq d_\text{local}$.} does not exceed the known step cost $c$, implemented via a squared hinge penalty that only activates when $d_\theta(o,o'){>}c$.

Intuitively, together with the quasimetric architecture of Appendix~\ref{appdx:arch}~(iv), which enforces the triangle inequality by construction, this formulation drives the model to behave like a \emph{cost-to-go function}: distances accumulate along paths, preserve local step costs, and cannot “shortcut” around the triangle inequality. Unlike explicit regression to ground-truth temporal distances, the QRL objective indirectly shapes the function through geometric constraints and value consistency, potentially yielding a more robust and transferable inductive structure for navigation.

\section{Ordinal Consistency: Qualitative Results}
\label{appdx:qualitative}

\paragraph{Procedure.}
For a set of held-out trajectories, we \emph{independently} compute predicted distances at each frame for every model under evaluation, using the last frame as the goal image.
For each trajectory, we form a time series of normalized distances per model and compute Kendall’s~$\tau$ against ground-truth temporal distance, yielding an ordinal consistency score per trajectory.
We additionally render qualitative plots overlaying (i) predicted distance curves, (ii) ground-truth distance (normalized), (iii) model confidence (when available), and (iv) the goal image.

\paragraph{Interpreting the qualitative trajectories.}
The examples show how distance predictions and auxiliary confidence change as goal-relevant cues appear. Overall progress can coexist with temporary increases or abrupt changes in predicted distance; ordinal consistency summarizes this ordering without requiring pointwise calibration.

\begin{figure*}[tbp]
\centering
\begin{subfigure}{0.95\linewidth}
  \centering
  \includegraphics[width=\linewidth]{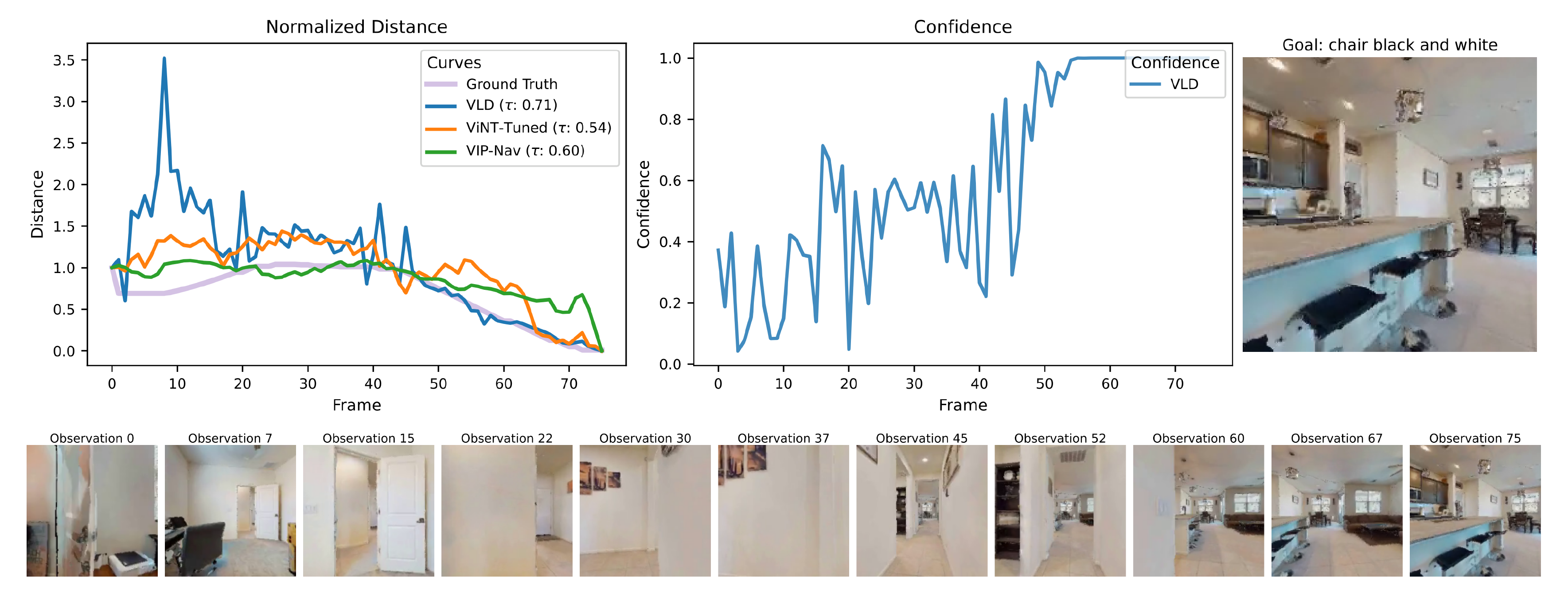}
  \caption{Habitat Example 1}
\end{subfigure}
\vspace{0.25cm}
\begin{subfigure}{0.95\linewidth}
  \centering
  \includegraphics[width=\linewidth]{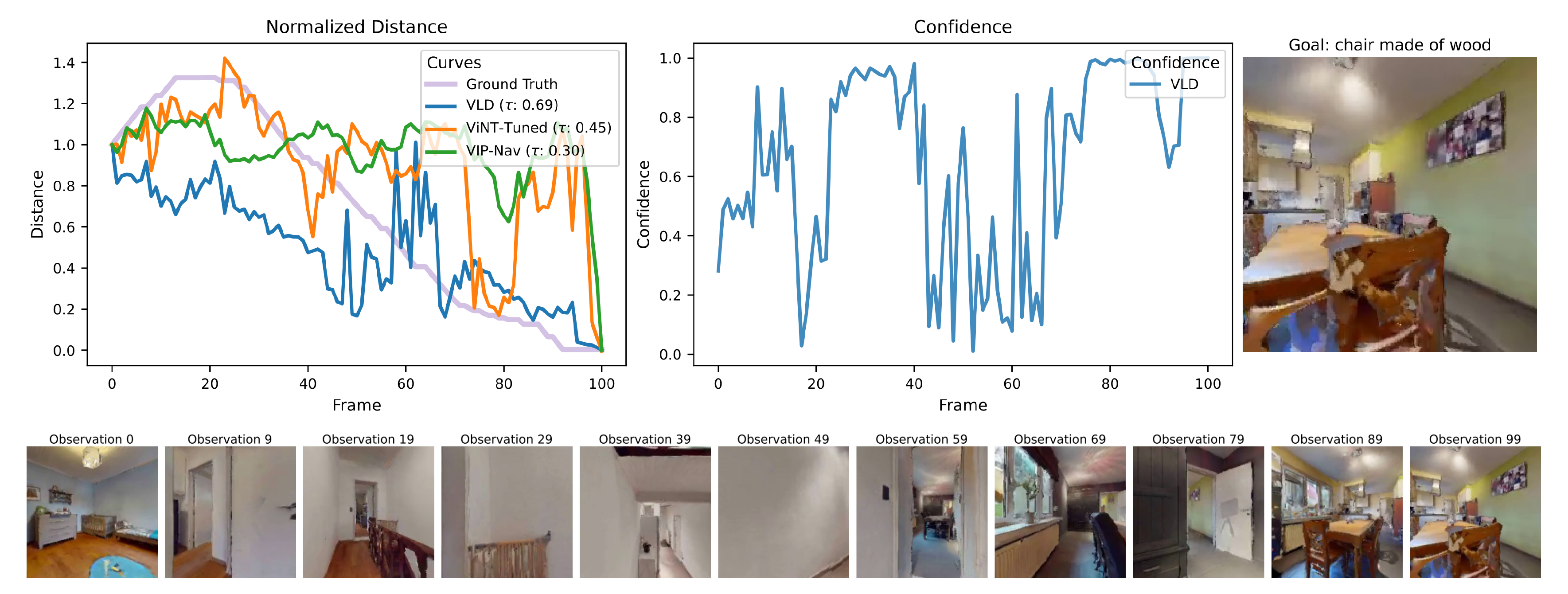}
  \caption{Habitat Example 2}
\end{subfigure}
\vspace{0.25cm}
\begin{subfigure}{0.95\linewidth}
  \centering
  \includegraphics[width=\linewidth]{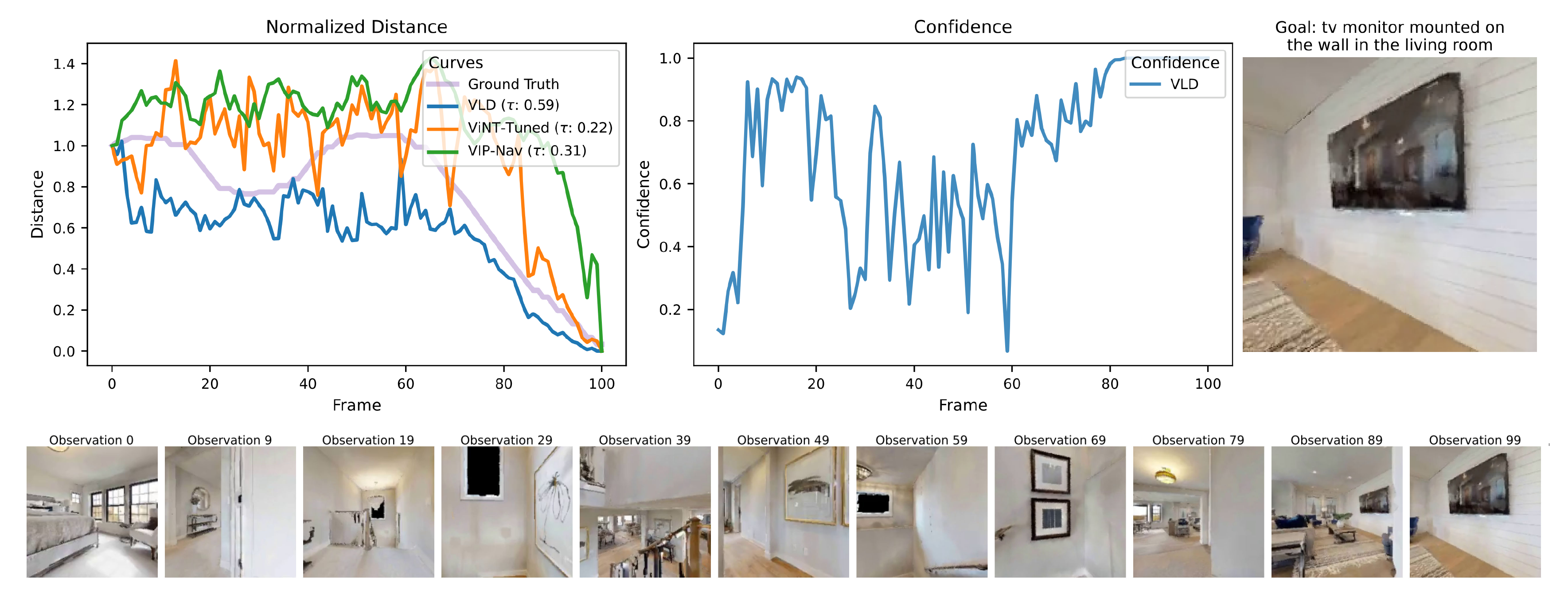}
  \caption{Habitat Example 3}
\end{subfigure}

\caption{\textbf{Habitat qualitative trajectories.} Each top row shows a full trajectory plot (normalized predicted distances, GT distance, and Kendall’s~$\tau$) with all models overlaid (VLD, ViNT-Tuned, VIP-Nav). Each bottom row displays the sequence of observations along the respective trajectory.}
\label{fig:qual_habitat}
\end{figure*}

\paragraph{Habitat examples (Figure~\ref{fig:qual_habitat}).}
VLD follows the broad decrease in temporal distance along the illustrated shortest-path-follower trajectories. The baseline curves show more deviations from this trend on these examples. Confidence typically increases when the goal enters the field of view.

\begin{figure*}[tbp]
\centering
\begin{subfigure}{0.95\linewidth}
  \centering
  \includegraphics[width=\linewidth]{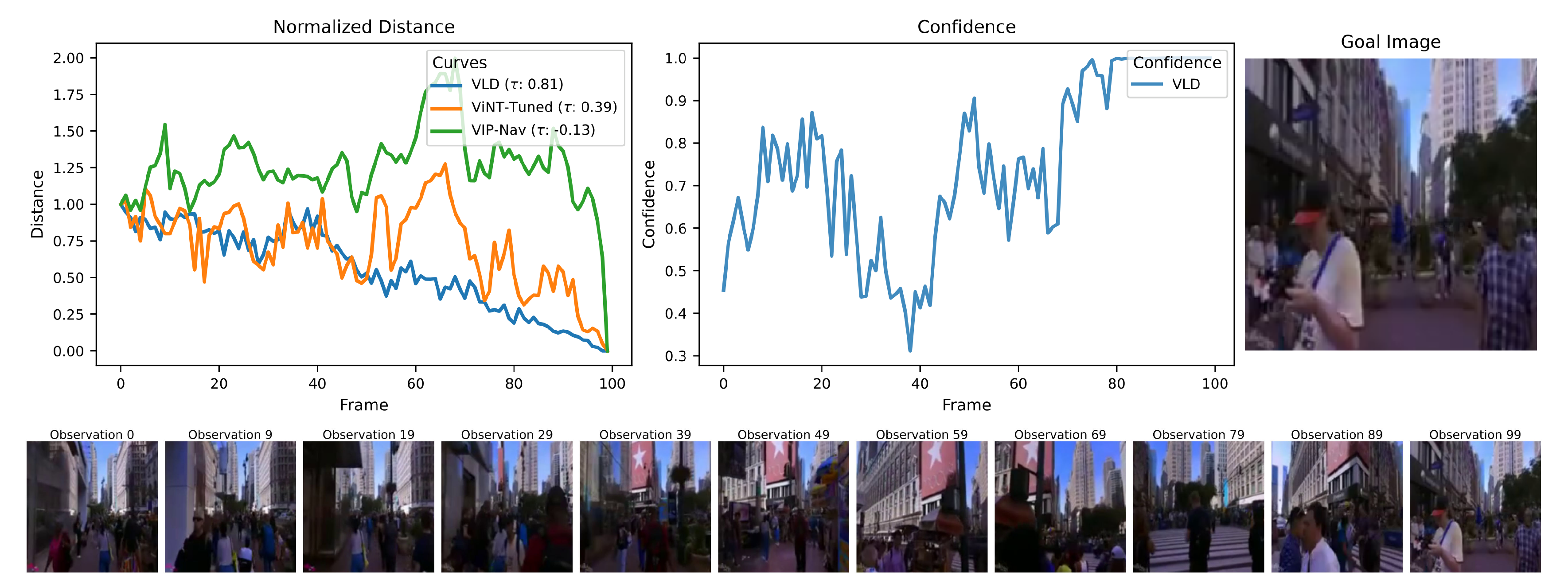}
  \caption{Real-world Example 1}
\end{subfigure}
\vspace{0.25cm}

\begin{subfigure}{0.95\linewidth}
  \centering
  \includegraphics[width=\linewidth]{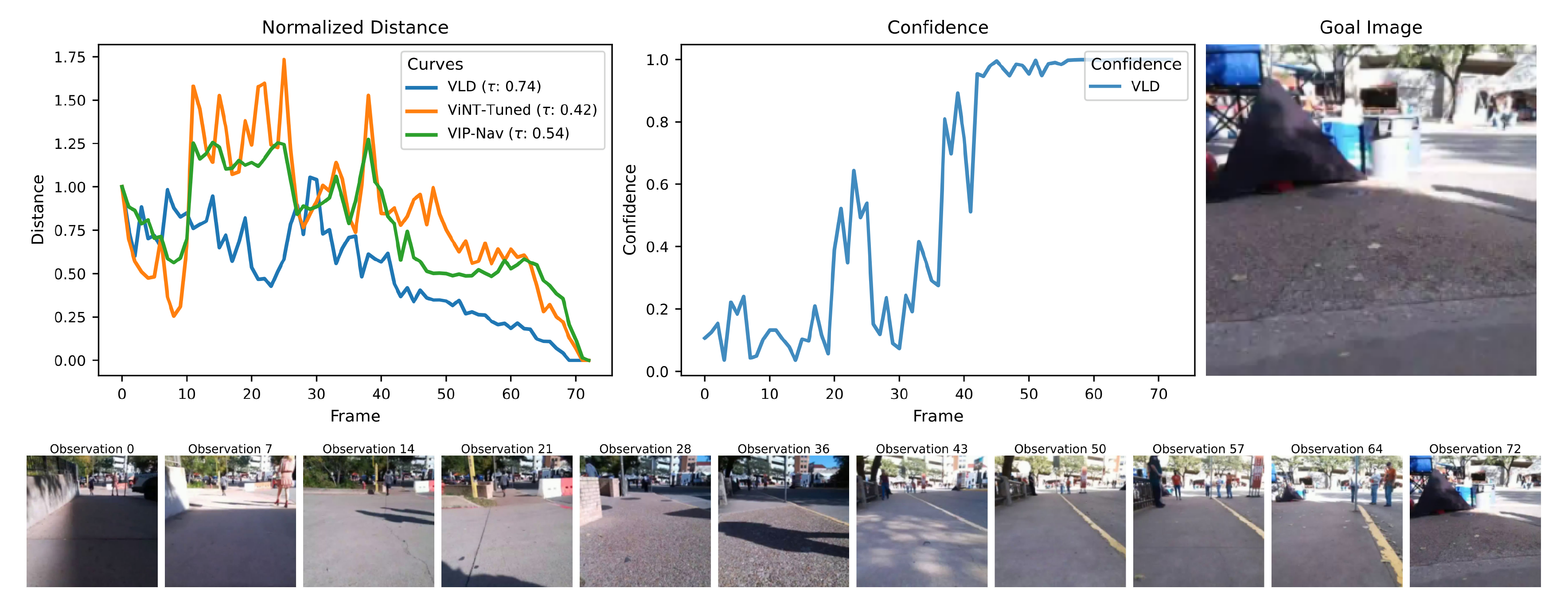}
  \caption{Real-world Example 2}
\end{subfigure}
\vspace{0.25cm}

\begin{subfigure}{0.95\linewidth}
  \centering
  \includegraphics[width=\linewidth]{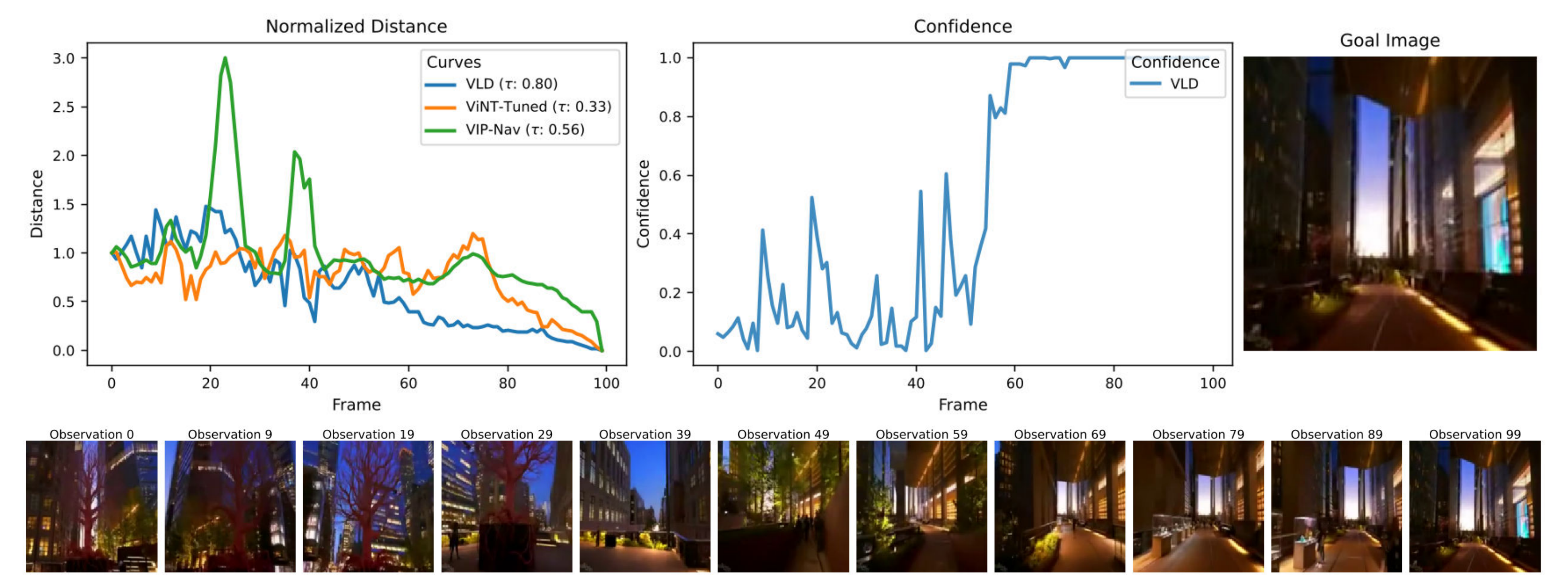}
  \caption{Real-world Example 3}
\end{subfigure}

\caption{\textbf{Real-world qualitative trajectories (Part 1).} The top rows of examples 1–3 show diverse “in-the-wild” and embodiment scenes with normalized predicted distances and Kendall’s $\tau$ overlaid for all models (VLD, ViNT-Tuned, VIP-Nav). Each bottom row displays the sequence of observations along the respective trajectory.}
\label{fig:qual_real_part1}
\end{figure*}

\begin{figure*}[tbp]
\centering
\begin{subfigure}{0.95\linewidth}
  \centering
  \includegraphics[width=\linewidth]{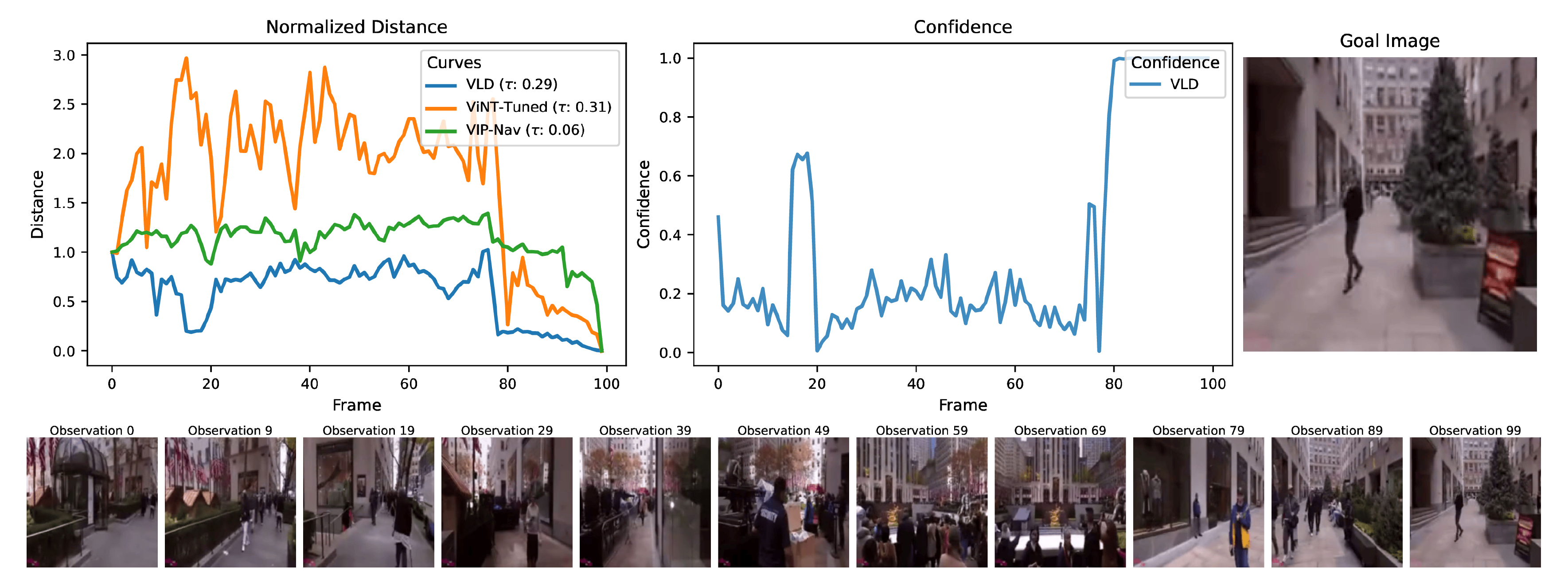}
  \caption{Real-world Example 4}
\end{subfigure}
\vspace{0.25cm}

\begin{subfigure}{0.95\linewidth}
  \centering
  \includegraphics[width=\linewidth]{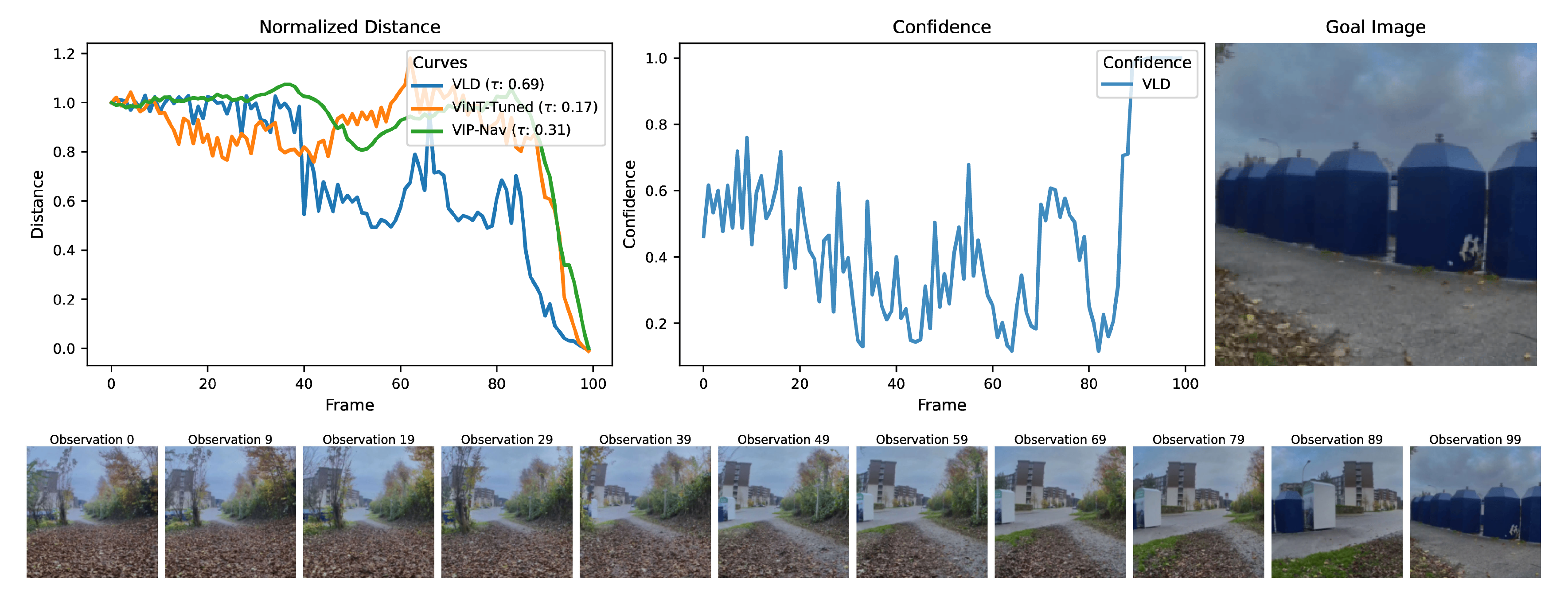}
  \caption{Real-world Example 5}
\end{subfigure}
\vspace{0.25cm}

\begin{subfigure}{0.95\linewidth}
  \centering
  \includegraphics[width=\linewidth]{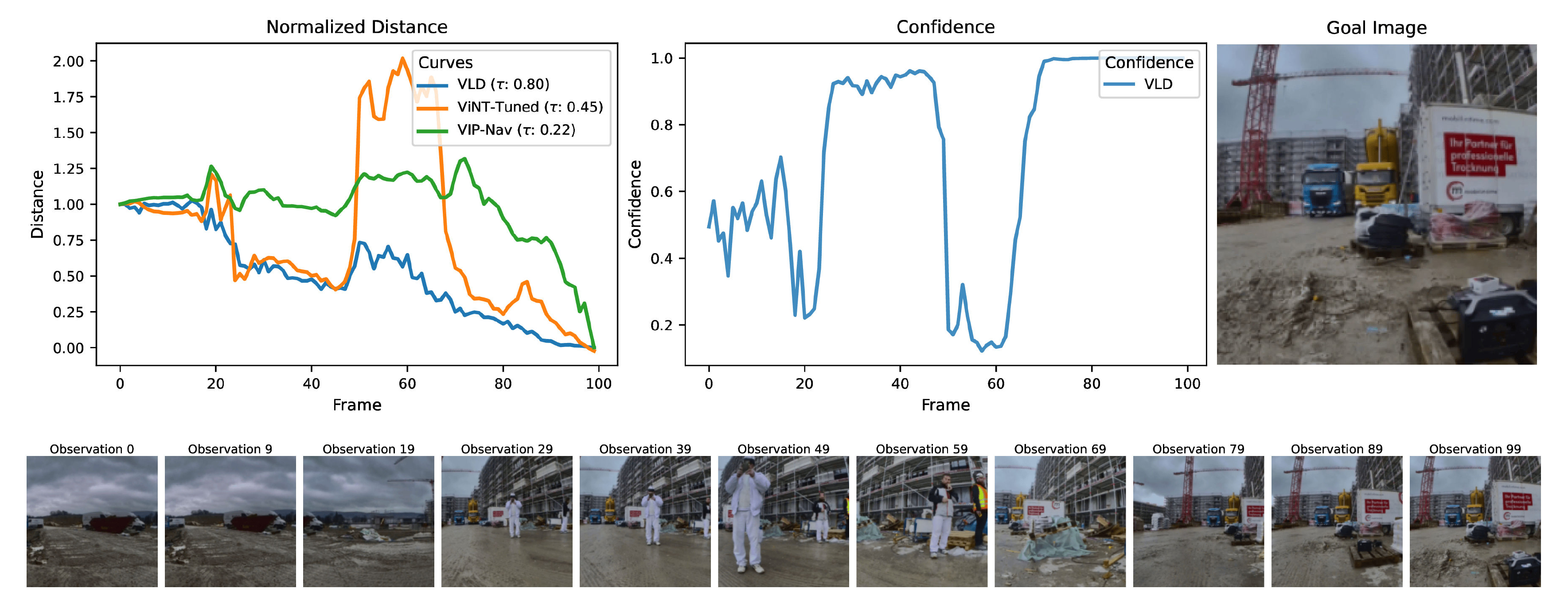}
  \caption{Real-world Example 6}
\end{subfigure}

\caption{\textbf{Real-world qualitative trajectories (Part 2).} Examples 4--6 show additional robot trajectories with normalized distance predictions and auxiliary confidence. Observation strips provide visual context for changes in the predictions. Together with Figure~\ref{fig:qual_real_part1}, these show both progress trends and local deviations on real imagery.}
\label{fig:qual_real_part2}
\end{figure*}

\paragraph{Real-world examples (Figures~\ref{fig:qual_real_part1}--\ref{fig:qual_real_part2}).}
The selected walking and robot trajectories contain viewpoint changes, temporary occlusions, and less regular motion than the synthetic examples. VLD generally preserves a decreasing trend despite local deviations caused by occlusions and viewpoint changes.

\begin{figure}[tbp]
\begin{center}
\includegraphics[width=0.9\linewidth]{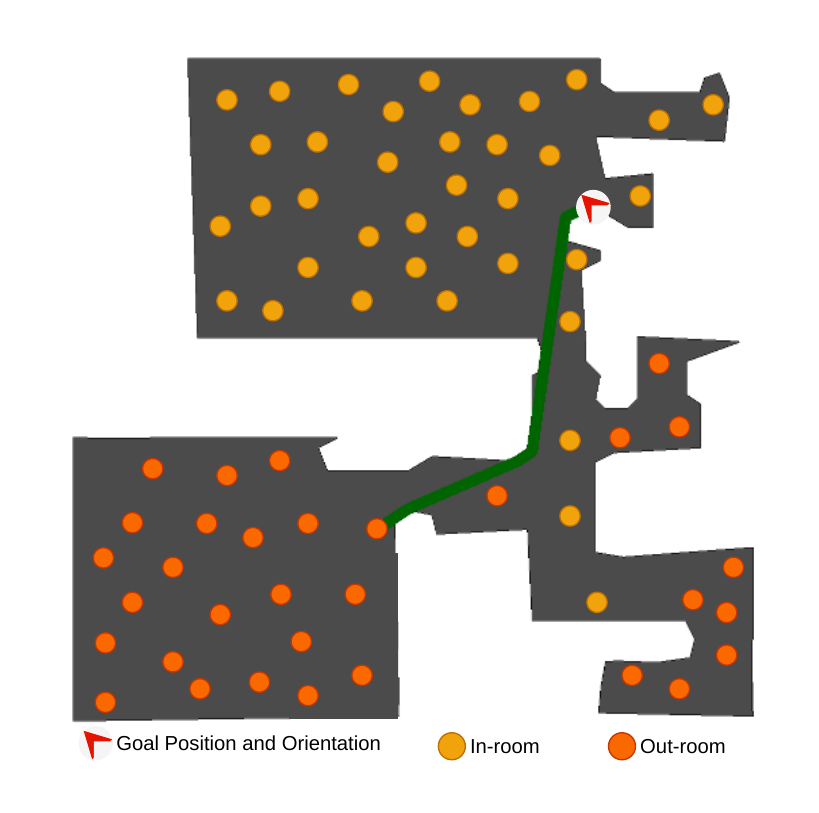}
\end{center}
\caption{\textbf{Evaluating distance accuracy via spatial point sampling.}
For each scene, we sample points both \emph{inside} and \emph{outside} the room containing the goal (red).
Following the criterion illustrated here, a point is considered \emph{in-room} if there exists at least one camera orientation at that location from which the goal image exhibits non-zero visual overlap; otherwise, it is labeled \emph{out-room}. \emph{A notable side effect of this definition is that certain hallway locations in this figure, adjacent to the goal room, may also be classified as ``in-room,'' simply because the goal becomes marginally visible from specific orientations at those positions.}%
}
\label{fig:dir_acc_map}
\end{figure}

\section{Distance Accuracy Evaluation}
\label{appdx:distance_accuracy}

To complement ordinal consistency and more directly assess whether VLD provides a meaningful navigation signal when the goal is visually out of view, we introduce a \emph{distance accuracy} metric that compares relative distances between pairs of observations.

For each validation episode in HM3D, we sample 100 viewpoints located in the same room as the goal, while explicitly selecting viewpoints and orientations such that the goal is \emph{not} visible. We also sample 100 viewpoints located outside that room (see Figure~\ref{fig:dir_acc_map}). Pairing these viewpoints yields three comparison categories: (i) \textbf{in-in} (both viewpoints in the goal room), (ii) \textbf{out-in} (one in the goal room, one outside), and (iii) \textbf{out-out} (both outside the goal room). For each category, we form all possible pairwise comparisons of predicted distance to the goal, and compute the fraction of pairs for which VLD correctly predicts which viewpoint is closer according to ground-truth geodesic distance.

Results are summarized in Table~\ref{tab:distance_accuracy}. Despite the absence of any direct visual overlap with the goal in all three categories, VLD identifies the closer viewpoint above the $50\%$ chance level in all three categories, with accuracy ranging from $53.32\%$ for out--out pairs to $65.48\%$ for out--in pairs.

We emphasize that this comparison is particularly challenging for two reasons. First, in most training videos, the agent intentionally moves toward the final goal, meaning that upon entering a room, they typically observe objects or structures that also appear in the goal image. This creates a strong visual bias that is not present in our evaluation setup. Second, the difficulty is compounded by the geometry of indoor environments.
Even when the agent stands in the same room as the goal image was captured, our evaluation forces it to consider orientations where the goal is not visible. In many realistic layouts, looking away from the goal often exposes features of other connected rooms—sometimes more prominently than the features of the room that actually contains the goal. For example, a goal image taken in a living room may face one direction, while the opposite orientation reveals an adjacent kitchen rather than the living room itself. As a result, the “same-room but opposite orientation’’ case can be surprisingly ambiguous and visually misleading, making the viewpoint-distance estimation task substantially harder than it may appear.

\begin{table}[tb]
  \centering
  \caption{Distance accuracy on HM3D validation scenes.
  The metric reports the fraction of pairwise comparisons in which VLD correctly
  identifies the closer viewpoint relative to the goal.}
  \label{tab:distance_accuracy}
    \footnotesize
  \begin{tabular}{@{}lc@{}}
    \toprule
    \textbf{Category} & \textbf{Accuracy (\%, ↑)} \\
    \midrule
    In--In (both in goal room)        & 58.61 \\
    Out--In (one in goal room)        & 65.48 \\
    Out--Out (neither in goal room)   & 53.32 \\
    \bottomrule
  \end{tabular}
\end{table}

\section{Distance Noise Models}
\label{appdx:ou_noise_abl}

\subsection{Ornstein-Uhlenbeck (OU) Distance Noise}
\label{appdx:ou_noie_descr}

\begin{figure*}[tbp]
\begin{center}
\includegraphics[width=0.95\linewidth]{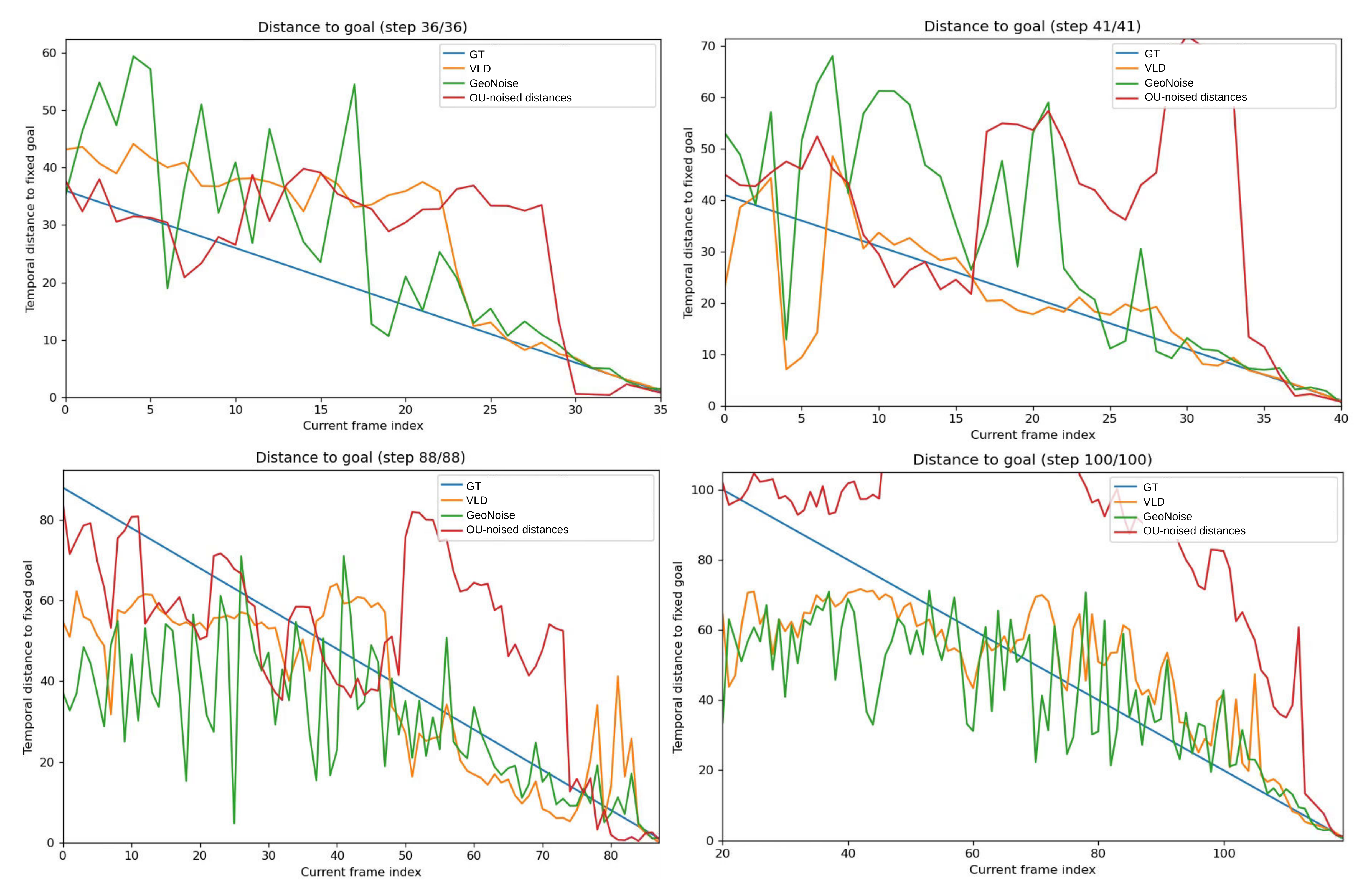}
\end{center}
\caption{Four example trajectories illustrating the behavior of different noise models.
\textcolor{mplC0}{Ground-truth (GT)} temporal distance (a linearly decreasing signal) is shown in \textcolor{mplC0}{blue}.
\textcolor{mplC1}{VLD} predictions are shown in \textcolor{mplC1}{orange}.
The \textcolor{mplC2}{GeoNoise}---the learned geometric-overlap noise model—produces the curves shown in \textcolor{mplC2}{green}, while
\textcolor{mplC3}{OU-noised distances} are shown in \textcolor{mplC3}{red}, obtained by applying an Ornstein–Uhlenbeck process directly to the privileged ground-truth distances.}
\label{fig:nose_comaprions}
\end{figure*}

Besides GeoNoise (Section~\ref{sec:noise_f}), we evaluate a generic noise model as a baseline. Since recurrent policies (LSTMs, GRUs) integrate observations over time~\cite{lstm, gru}, we use temporally correlated noise based on an \emph{Ornstein–Uhlenbeck (OU) process}, a type of “red noise” widely used in RL exploration~\cite{ddpg, noise_test}. For each episode, we maintain a noise state updated as:
\[
\epsilon_{t}= \alpha \,\epsilon_{t-1} + \sigma \, \xi_{t},
\quad \xi_t \sim \mathcal{N}(0,1),
\]
where $\alpha=0.9$ and $\sigma=0.1$ are fixed hyperparameters controlling correlation and magnitude. To mimic unpredictable prediction failures, we additionally inject occasional spike perturbations. Concretely, at each timestep we sample a Bernoulli indicator $b_t \in \{0,1\}$ with probability $0.05$ of being active, and—conditioned on a spike occurring—a Gaussian amplitude $\Psi_t \sim \mathcal{N}(0,4)$. The final noise is therefore
\[
\epsilon'_{t} = \epsilon_t \;+\; b_t\,\Psi_t,
\qquad
b_t \sim \mathrm{Bernoulli}(0.05),\,
\Psi_t \sim \mathcal{N}(0,4).
\]
For more realistic behavior, the noise signal is applied to the ground-truth \emph{geometric distance-to-goal} ($d$) provided by the simulator, conditioned as follows:
\[
\hat d = \max\!\big(0,\; d + \epsilon'_t\big),
\qquad |\epsilon'_t| \le f(d)=\exp\!\big(\sqrt{d}\big) - 1,
\]
where $f(d)$ increases with $d$, reducing the admissible noise magnitude near the goal. This prevents unrealistic negative distances or excessively large errors when the agent is close to the target.

For the confidence-input variant, we derive an auxiliary confidence value from the injected noise magnitude. We first compute the proxy:
\[
c' = \exp(-\kappa|\epsilon'_t|),
\]
where $\kappa$ is a scaling constant, and then perturb $c'$ with an independent OU process ($\phi$) to avoid trivially revealing the injected noise to the policy:
\[
\hat c = \text{clip}(c' + \phi_t,\, 0,\, 1).
\]
The distance-only OU policy receives $\hat d$; the confidence-input variant receives the pair $(\hat d,\hat c)$.

\subsection{Noise Ablation Results}
Table~\ref{tab:vld_policy_transfer} in the main text compares GeoNoise and OU training on HM3D and Gibson. With their own training signals, OU-trained policies achieve higher success than GeoNoise-trained ones ($94.4\%$ and $94.5\%$ SR on HM3D and Gibson without confidence input, $95.6\%$ and $97.3\%$ with it). After substitution by VLD, these scores fall to $4.9\%$ and $11.5\%$ ($16.5\%$ and $47.1\%$ with confidence input), while the corresponding GeoNoise-trained policies retain $58.7\%$ and $76.2\%$ ($52.1\%$ and $68.2\%$). High success with the training input alone is therefore not a reliable indicator of compatibility with learned visual distances. Figure~\ref{fig:nose_comaprions} illustrates the difference: OU noise perturbs the privileged distance independently of the scene, whereas GeoNoise reproduces the viewpoint-dependent structure of VLD predictions. Confidence input helps the OU-trained policy after substitution but does not improve the GeoNoise-trained policy, which motivates the distance-only main configuration. The qualitative rollouts (Appendix~\ref{sec:qual_nav}) show that GeoNoise-trained policies perform periodic scans that expose the agent to different goal cues, a behavior suited to viewpoint-dependent predictions.

\section{Navigation Policy: Qualitative Results}
\label{sec:qual_nav}

We visualize representative rollouts from the navigation policy when deployed with VLD.
These examples illustrate how the agent (i) explores previously unseen spaces, (ii) uses VLD feedback to prioritize promising directions, (iii) detects partial visual cues from the goal, and (iv) converges to the target.
We also show characteristic failure modes.

\subsection{Successful Navigation}
\label{sec:qual_succ_nav}

\begin{figure*}[tbp]
\centering

\begin{subfigure}{0.95\linewidth}
  \centering
  \includegraphics[width=\linewidth]{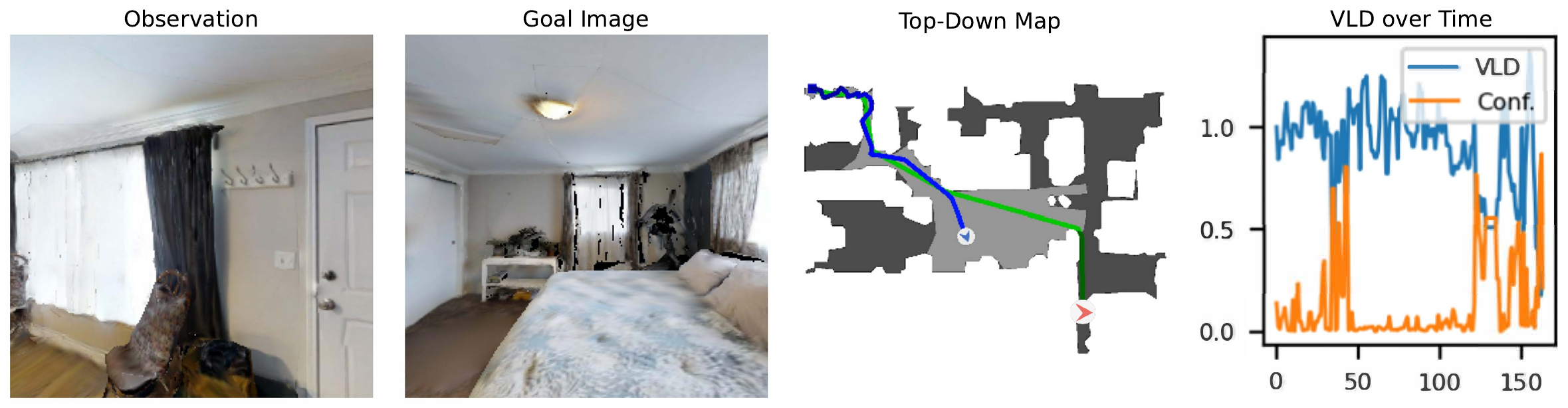}
  \caption{The agent enters a new space with no immediate cues toward the goal.}
\end{subfigure}

\begin{subfigure}{0.95\linewidth}
  \centering
  \includegraphics[width=\linewidth]{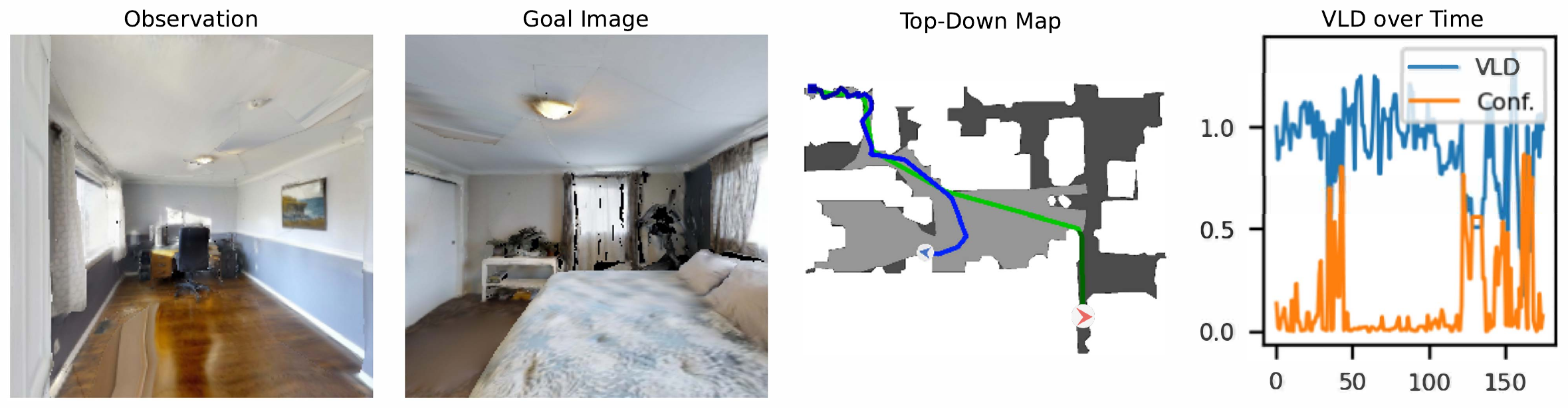}
  \caption{It performs an initial sweep, collecting VLD predictions across multiple headings.}
\end{subfigure}

\begin{subfigure}{0.95\linewidth}
  \centering
  \includegraphics[width=\linewidth]{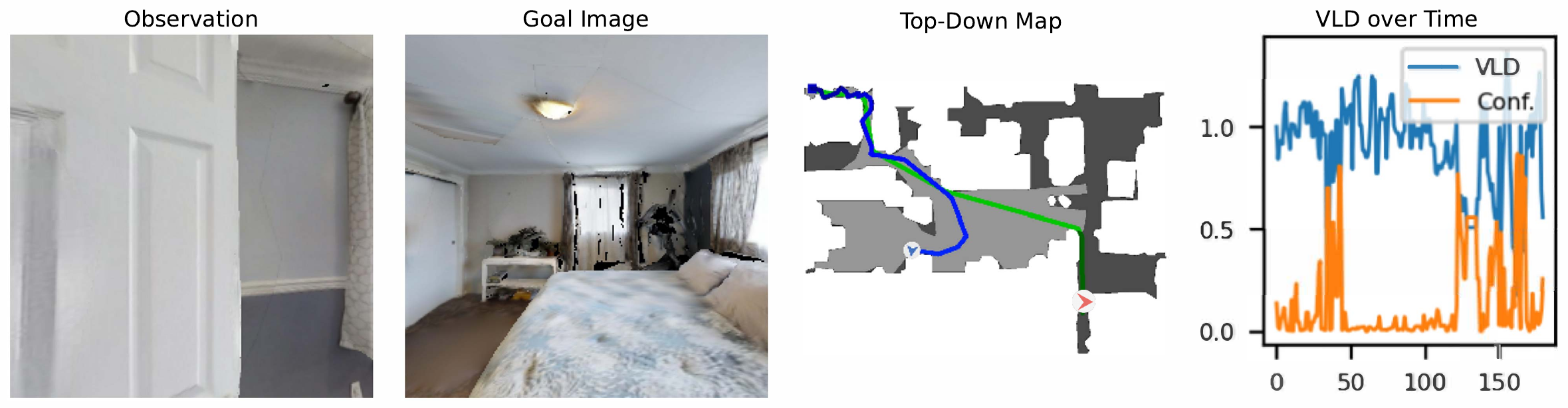}
  \caption{A full rotation provides a broader set of viewpoints for comparing predicted distances.}
\end{subfigure}

\begin{subfigure}{0.95\linewidth}
  \centering
  \includegraphics[width=\linewidth]{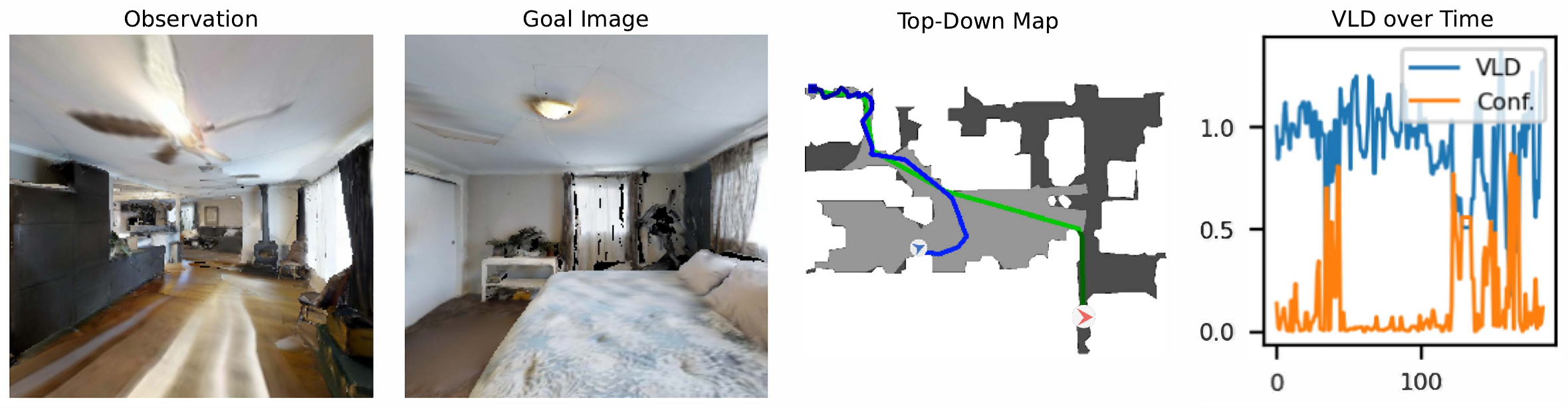}
  \caption{The agent commits to the direction with the lowest VLD-predicted distance and continues moving.}
\end{subfigure}

\caption{\textbf{Exploration driven by VLD.}
The policy uses VLD as a scalar navigation signal, rotating and evaluating multiple viewpoints before selecting an exploration direction.
Each step shows: (i) the agent’s current observation (blue arrow) and the goal position and orientation (red arrow), (ii) the goal image, (iii) a top-down map with the taken trajectory in blue (green shows the optimal path), and (iv) the plot of VLD-predicted distance and confidence over past steps.}
\label{fig:exploration_nav}
\end{figure*}

\paragraph{Exploration behavior.}
When entering an unfamiliar room, the agent typically performs a brief 360-degree scan—often multiple partial rotations—before committing to a direction.
VLD encourages movement toward headings with lower predicted distances: as the agent rotates, it evaluates several candidate viewpoints, compares predicted distances, and then proceeds along the direction with the minimum value.
See Figure~\ref{fig:exploration_nav} for an illustration.

\begin{figure*}[tbp]
\centering

\begin{subfigure}{0.95\linewidth}
  \centering
  \includegraphics[width=\linewidth]{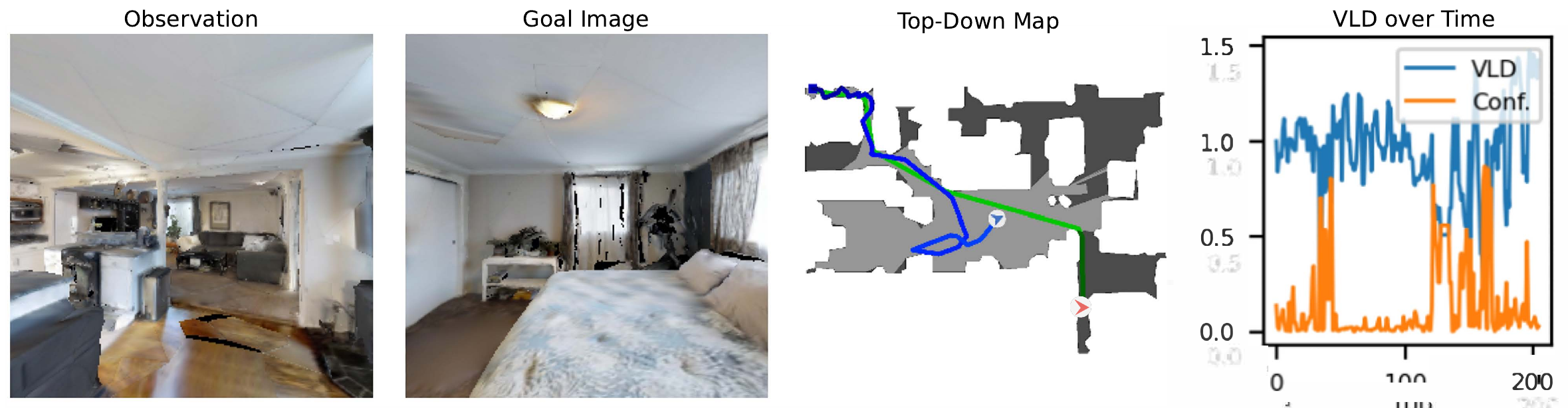}
  \caption{The agent continues exploring, guided by VLD predictions.}
\end{subfigure}

\begin{subfigure}{0.95\linewidth}
  \centering
  \includegraphics[width=\linewidth]{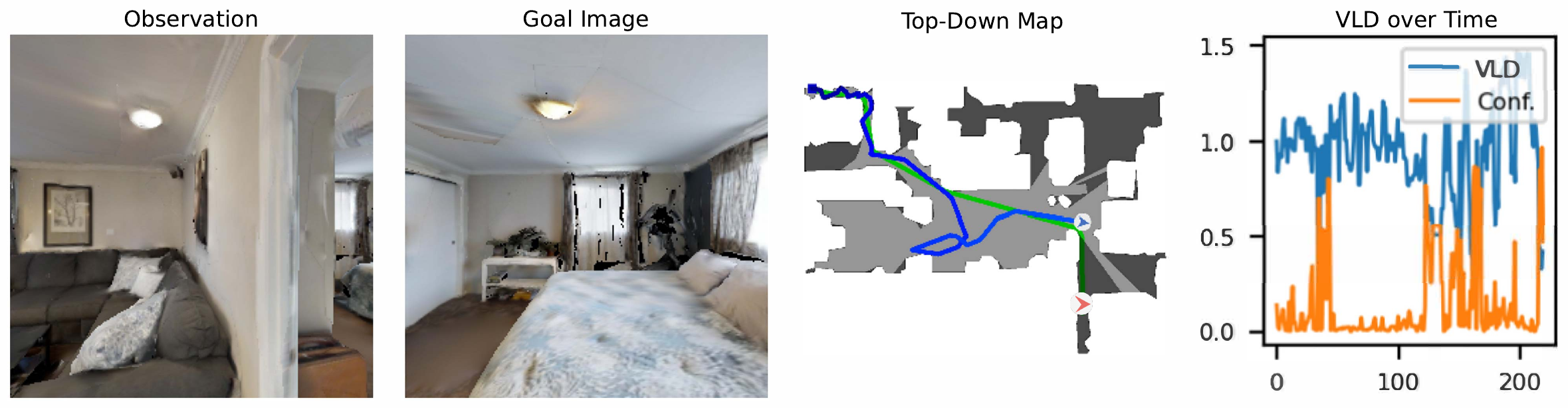}
  \caption{First partial cue from the goal appears; VLD confidence spikes and predicted distance drops.}
\end{subfigure}

\begin{subfigure}{0.95\linewidth}
  \centering
  \includegraphics[width=\linewidth]{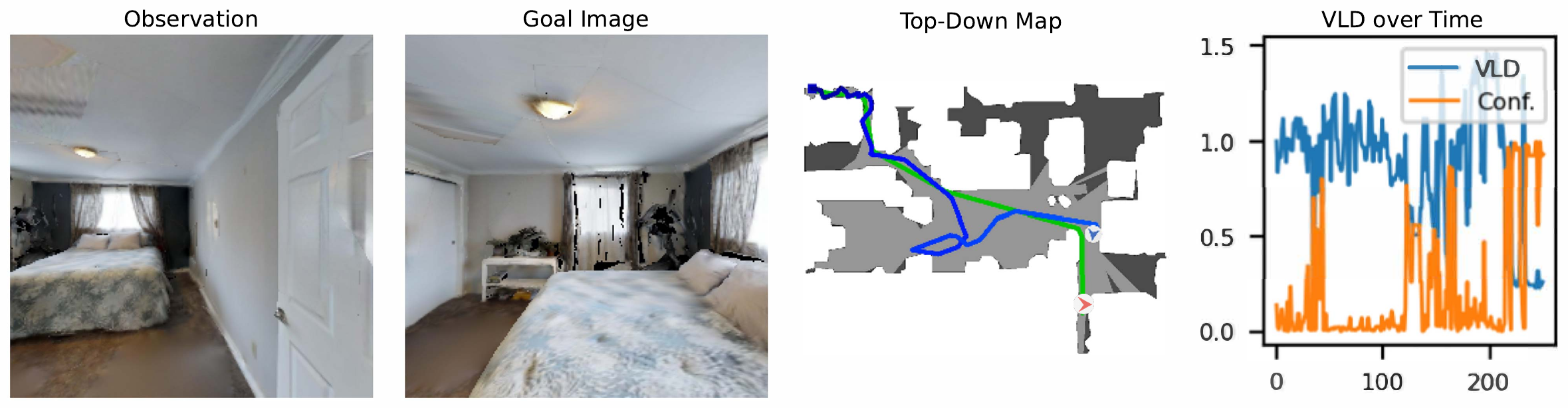}
  \caption{The agent commits to the identified direction; distance decreases sharply.}
\end{subfigure}

\begin{subfigure}{0.95\linewidth}
  \centering
  \includegraphics[width=\linewidth]{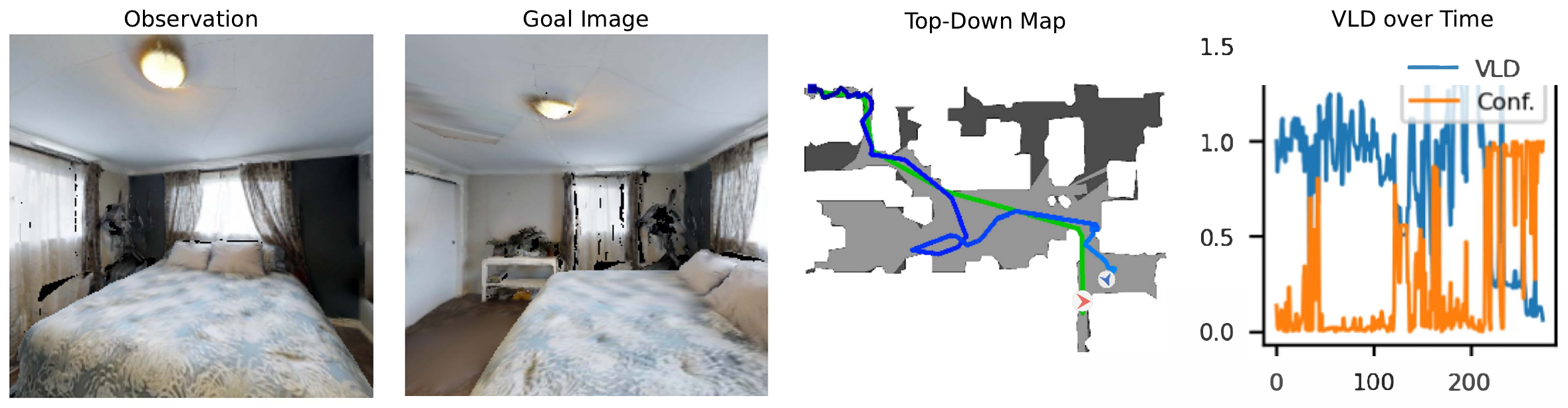}
  \caption{The goal object is reached and the episode terminates successfully.}
\end{subfigure}

\caption{\textbf{Successful navigation.}
Once partial visual overlap occurs, predicted distance decreases and the policy approaches the goal. The confidence trace is shown for analysis.
Each step shows: (i) the agent’s current observation  (blue arrow) and the goal position and orientation (red arrow), (ii) the provided goal image, (iii) a top-down map with executed (blue) and optimal trajectories (green), and (iv) VLD distance and confidence plots.}
\label{fig:success_nav}
\end{figure*}

\paragraph{Detecting and approaching the goal.}
In successful rollouts, the agent eventually encounters partial visual overlap with the goal image.
At this moment, VLD confidence sharply increases while the predicted distance drops, coinciding with the policy's turn toward the goal. The confidence trace is shown for analysis only; the policy receives only the distance.
Figure~\ref{fig:success_nav} shows a sequence illustrating this behavior.

\subsection{Failure Cases}
\label{sec:qual_fail_nav}

To better characterize how and why navigation attempts fail, we categorize the unsuccessful episodes of the VLD-driven policy on both benchmarks into three primary modes (Table~\ref{tab:failure_modes}).
Time-outs are the most common failure on both HM3D (71\%) and Gibson (62\%), followed by near-match failures (28\% and 29\%). Bad or uninformative goal images account for only 1\% of the failures on HM3D but 9\% on Gibson, consistent with the Gibson task design discussed below. Long-horizon exploration is the main challenge: on Gibson episodes whose shortest path requires at least $50$ steps, the VLD-driven policy reaches $53\%$ SR.

\begin{table}[tbp]
\centering
\footnotesize
\caption{Distribution of navigation failure modes of the VLD-driven policy, as percentages of unsuccessful episodes.}
\label{tab:failure_modes}
\begin{tabular}{lcc}
\toprule
\textbf{Failure Type} & \textbf{HM3D (\%)} & \textbf{Gibson (\%)} \\
\midrule
Time-out failure           & 71 & 62 \\
Near-match failure         & 28 & 29 \\
Bad or uninformative goal  & 1 & 9 \\
\bottomrule
\end{tabular}
\end{table}

\begin{figure*}[tbp]
\centering

\begin{subfigure}{0.95\linewidth}
  \centering
  \includegraphics[width=\linewidth]{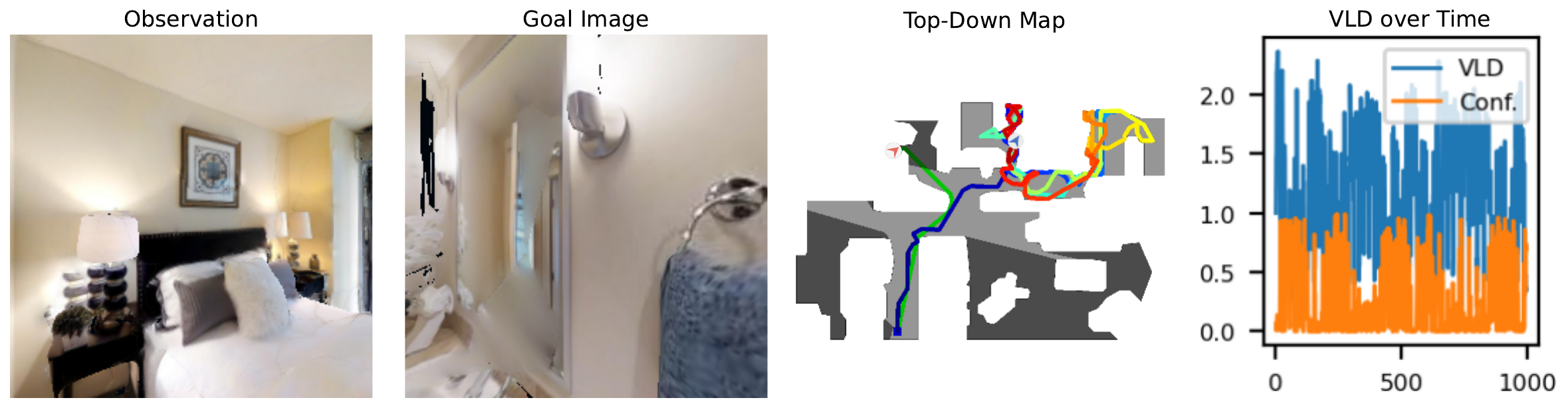}
  \caption{Time-out: the agent explores but never finds a direction with sufficiently decreasing VLD distance.}
  \label{fig:time_out}
\end{subfigure}

\begin{subfigure}{0.95\linewidth}
  \centering
  \includegraphics[width=\linewidth]{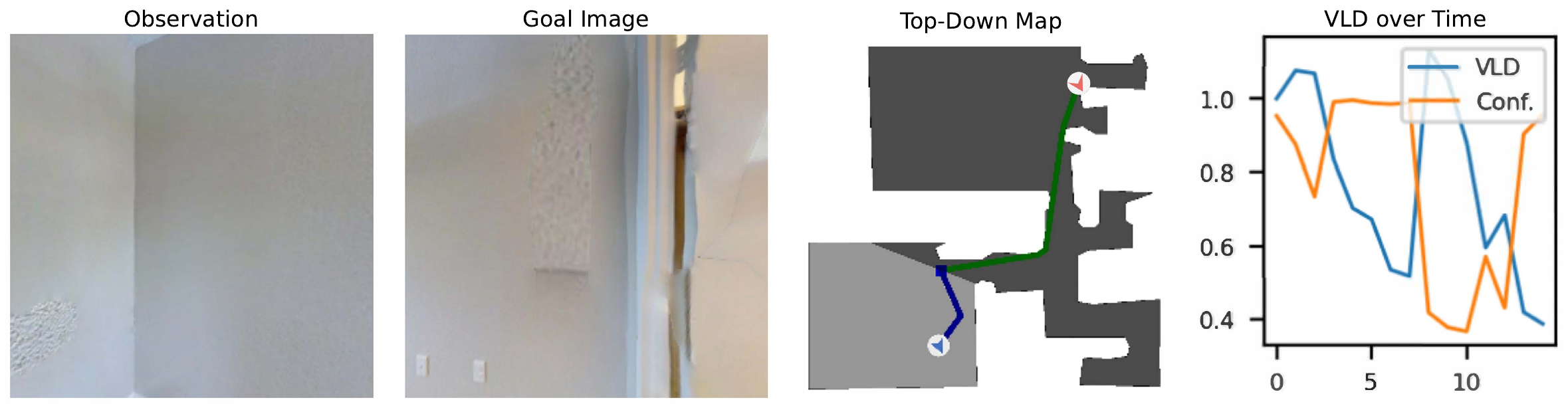}
  \caption{Ambiguous goal image induces misalignment between the intended and perceived target.}
  \label{fig:bad_goal}
\end{subfigure}

\begin{subfigure}{0.95\linewidth}
  \centering
  \includegraphics[width=\linewidth]{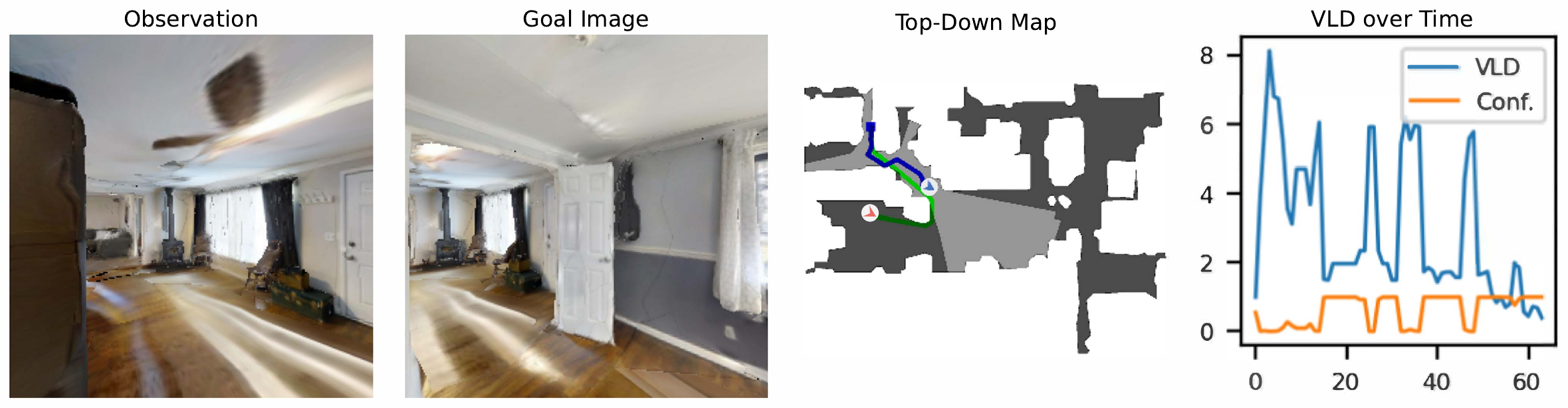}
  \caption{Near-match failure: the observation looks deceptively similar to the goal image, but the true goal is physically distant.}
  \label{fig:near-match1}
\end{subfigure}

\begin{subfigure}{0.95\linewidth}
  \centering
  \includegraphics[width=\linewidth]{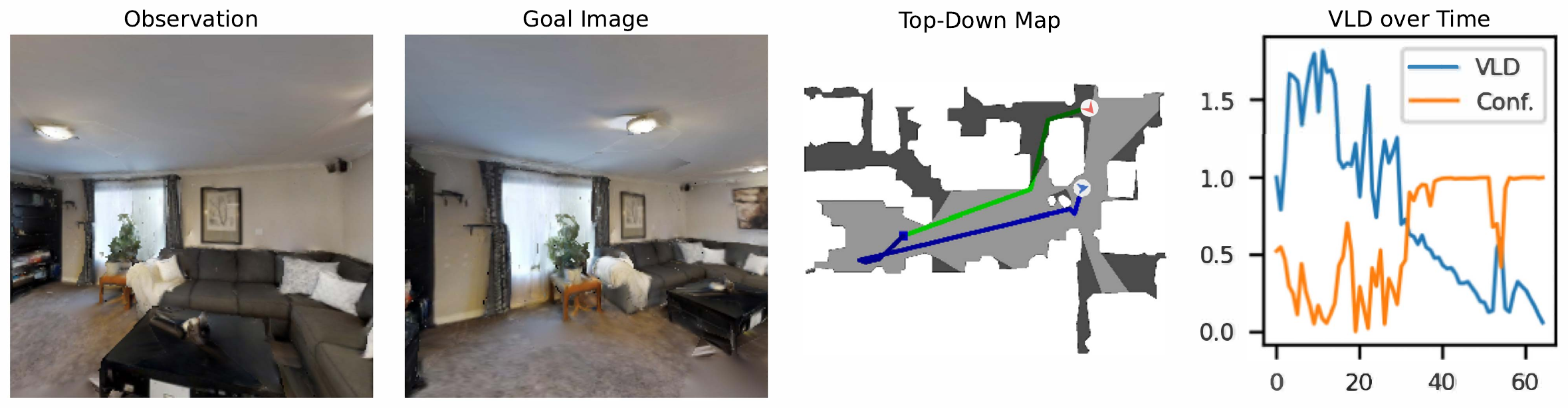}
  \caption{Another near-match failure: visual similarity misleads the policy despite incorrect spatial alignment.}
  \label{fig:near-match2}
\end{subfigure}

\caption{\textbf{Failure cases.}
Failures arise from long-horizon exploration without explicit memory (time-outs), visually deceptive near-matches, or ambiguous goal definitions.
Each example shows: (i) the agent’s current observation  (blue arrow) and the goal position and orientation (red arrow), (ii) the goal image, (iii) the top-down map with the executed (gradient color from dark blue at the start, to red color at the steps close to max steps limit) and optimal paths (green), and (iv) the VLD distance/confidence trace.}
\label{fig:failure_nav}
\end{figure*}

We visualize representative examples of these failure modes in Figure~\ref{fig:failure_nav}.

\paragraph{(1) Time-out failures.}
Most failures arise from trajectories where the agent becomes stuck exploring visually similar but spatially incorrect regions.
Equipping the agent with explicit memory—either short- or long-term—of previously visited locations, for example via a topological map or episodic graph~\cite{vint, feng2025imagegoalnavigationusingrefined}, may help mitigate such failures (see Figure~\ref{fig:time_out}).
However, memory augmentation is outside the scope of this work, as our focus is on evaluating VLD as a standalone navigation signal.

\paragraph{(2) Near-match failures.}
A second failure mode originates from the properties of the distance function itself.
In these \emph{near-match failures}, the agent encounters a view that appears deceptively similar to the goal image even though the true goal is still physically distant (Figures~\ref{fig:near-match1}–\ref{fig:near-match2}).  However, because the policy receives goal information through a scalar distance signal—which is naturally biased toward feature overlap—the agent may incorrectly conclude that it has reached the goal and stop prematurely.

In some near-match cases, the target object is visible from a different perspective, but the agent has not reached the goal location required by the benchmark, so these episodes count as failures.

\paragraph{(3) Bad or uninformative goal images.}
A third failure class stems from task design in the Gibson environment.
As goal viewpoints are sampled from random orientations, some goal images are highly ambiguous or visually uninformative (Figure~\ref{fig:bad_goal}).
In these cases, a predictor may struggle, as the target may not correspond to a distinctive or semantically meaningful observation.

\section{Real-World Experiments}
\label{sec:real-wolrd-expr}

\subsection{Experimental Setup}

We evaluate VLD + GeoNoise and FGPrompt-EF on a physical TurtleBot4-Light robot~\cite{turtlebot4_manual} equipped with a Logitech MX Brio 705 RGB camera with a $90^\circ$ field of view. The camera is mounted approximately $1\,\text{m}$ above the ground to match the RL training setup. Compass and GPS observations required by the policies are estimated from the robot's odometry.

\begin{figure*}[tb]
\centering
\includegraphics[width=\linewidth]{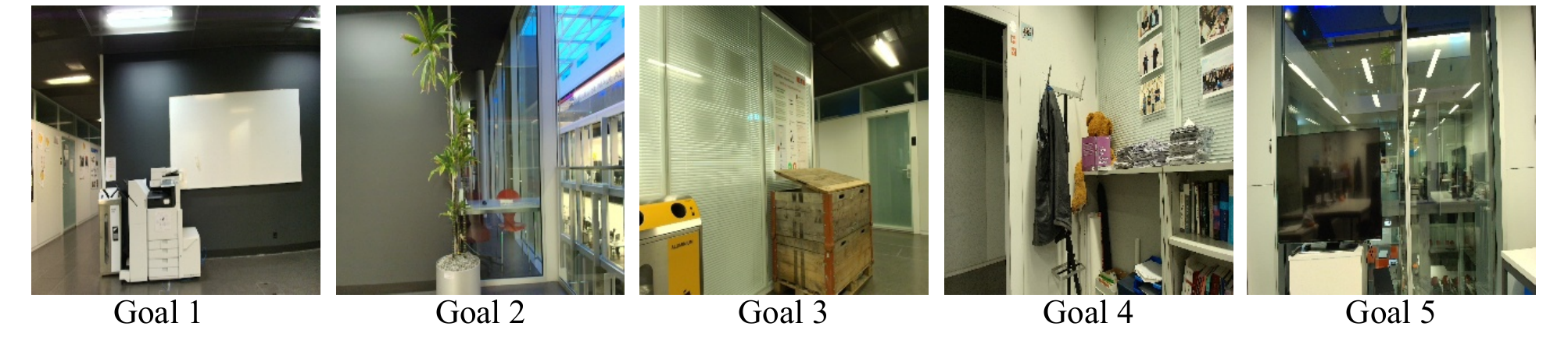}
\caption{\textbf{Goal images used in the real-world navigation experiments.}}
\label{fig:real-goals}
\end{figure*}

We define five goal locations, each represented by an image captured at the target position (Figure~\ref{fig:real-goals}). The first task starts at a fixed position $S$ from which $G_1$ is not visible. Each subsequent task starts at the previous goal location: $G_1$ for the task targeting $G_2$, and so on. We run \textbf{five independent repetitions per goal and per method}, giving $25$ navigation attempts for each method. Both methods are transferred without real-world fine-tuning. Each attempt is limited to $500$ actions, turn actions rotate the robot by approximately $30^\circ$, and after a failed attempt the robot restarts from the start position of the same task.

The tasks cover different exploration demands. The goal scene becomes visible after a small rotation or short movement for $G_1\rightarrow G_2$. Reaching $G_1$ from $S$, or $G_5$ from $G_4$, requires exploring in the appropriate direction before the goal appears. The routes $G_2\rightarrow G_3$ and $G_3\rightarrow G_4$ require navigation across multiple spaces or rooms.

For context, we also record trajectories from a human operator who knows the environment and the goal locations; these require no exploration and provide a reference action count. Each forward action moves the robot approximately $0.5\,\text{m}$, as in simulation.

\subsection{Results}

Table~\ref{tab:real_world_results} reports success rate (SR), computed as successful attempts divided by all attempts, and the average number of actions taken in \emph{successful trials only}. The overall SR pools all $25$ attempts per method. Since the methods succeed on different trials, action counts should be read together with SR.

\begin{table}[tb]
  \centering
  \small
  \caption{\textbf{Real-world navigation results.} Five repetitions at each of five goals give $25$ attempts per method. Average steps count actions in successful trials only. Human action counts provide a reference without exploration. Bold indicates the best SR, including ties.}
  \label{tab:real_world_results}
  \begin{tabular}{@{}lrrrrrr@{}}
    \toprule
    \textbf{Method} & \textbf{G1} & \textbf{G2} & \textbf{G3} & \textbf{G4} & \textbf{G5} & \textbf{Overall} \\
    \midrule
    \multicolumn{7}{l}{\textit{Success rate (SR $\uparrow$)}} \\
    VLD + GeoNoise & \textbf{1.00} & \textbf{0.80} & \textbf{0.80} & 0.80 & \textbf{1.00} & \textbf{0.88} \\
    FGPrompt-EF & 0.00 & 0.40 & 0.60 & \textbf{1.00} & \textbf{1.00} & 0.60 \\
    \midrule
    \multicolumn{7}{l}{\textit{Average steps over successful trials}} \\
    VLD + GeoNoise & 58.6 & 138.4 & 298.6 & 342.6 & 80.8 & -- \\
    FGPrompt-EF & -- & 326.0 & 271.6 & 75.4 & 106.4 & -- \\
    \midrule
    Human (reference steps) & 29 & 14 & 36 & 28 & 13 & -- \\
    \bottomrule
  \end{tabular}
\end{table}

VLD + GeoNoise reaches every goal in at least four of five trials, succeeding in $22/25$ attempts ($88\%$). FGPrompt-EF succeeds in $15/25$ attempts ($60\%$).

The VLD-driven policy periodically performs $360^\circ$ scans before committing to forward motion, as it does in simulation. These rotations add actions, but allow the policy to reassess the distance signal across viewpoints and detect weak goal cues. The qualitative trajectories suggest that this behavior helps the agent find promising directions when the goal is initially out of view.

FGPrompt-EF often selects a direction early and explores alternative directions less frequently. When the initial choice is incorrect, it can struggle to recover. For $G_1$, it does not succeed in any of five trials and frequently issues a \textit{stop} action far from the target without visible goal cues. We hypothesize that sim-to-real discrepancies in perception or stopping decisions contribute to this behavior.

\begin{figure*}[t]
\centering

\begin{subfigure}{0.82\linewidth}
  \centering
  \includegraphics[width=\linewidth]{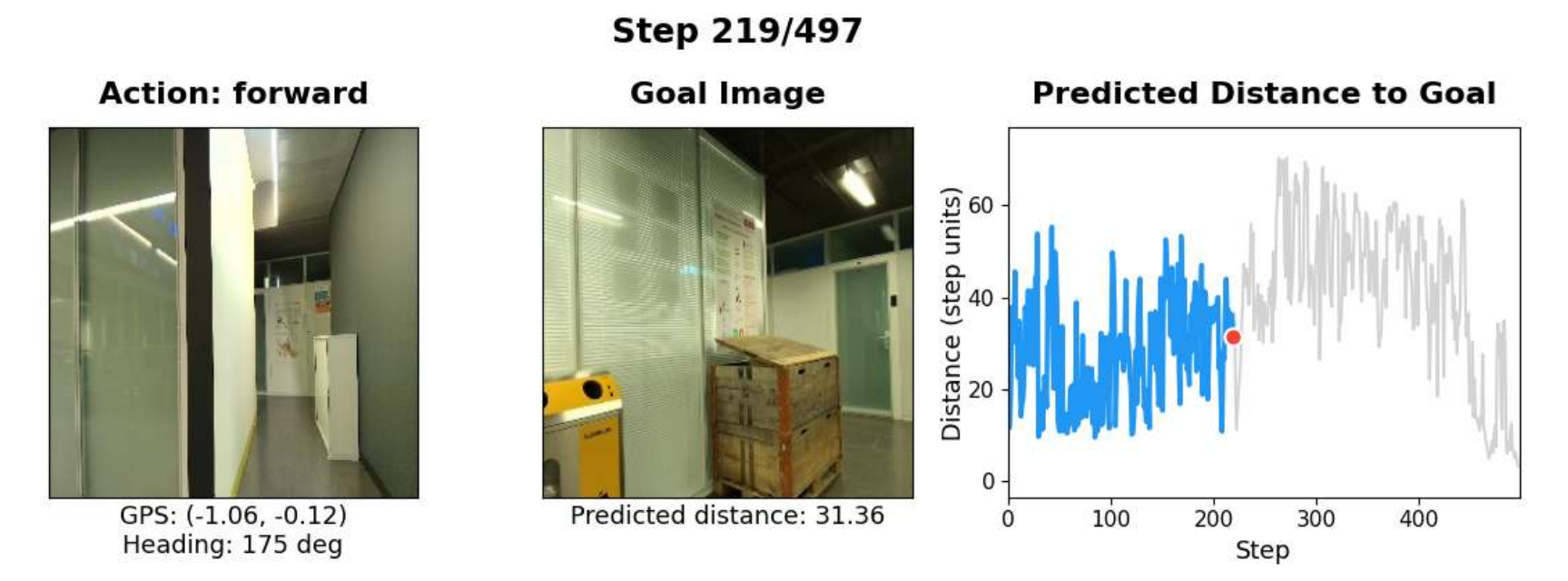}
  \caption{After initial exploration with no visible goal cues, the agent chooses to explore the hallway.}
\end{subfigure}

\begin{subfigure}{0.82\linewidth}
  \centering
  \includegraphics[width=\linewidth]{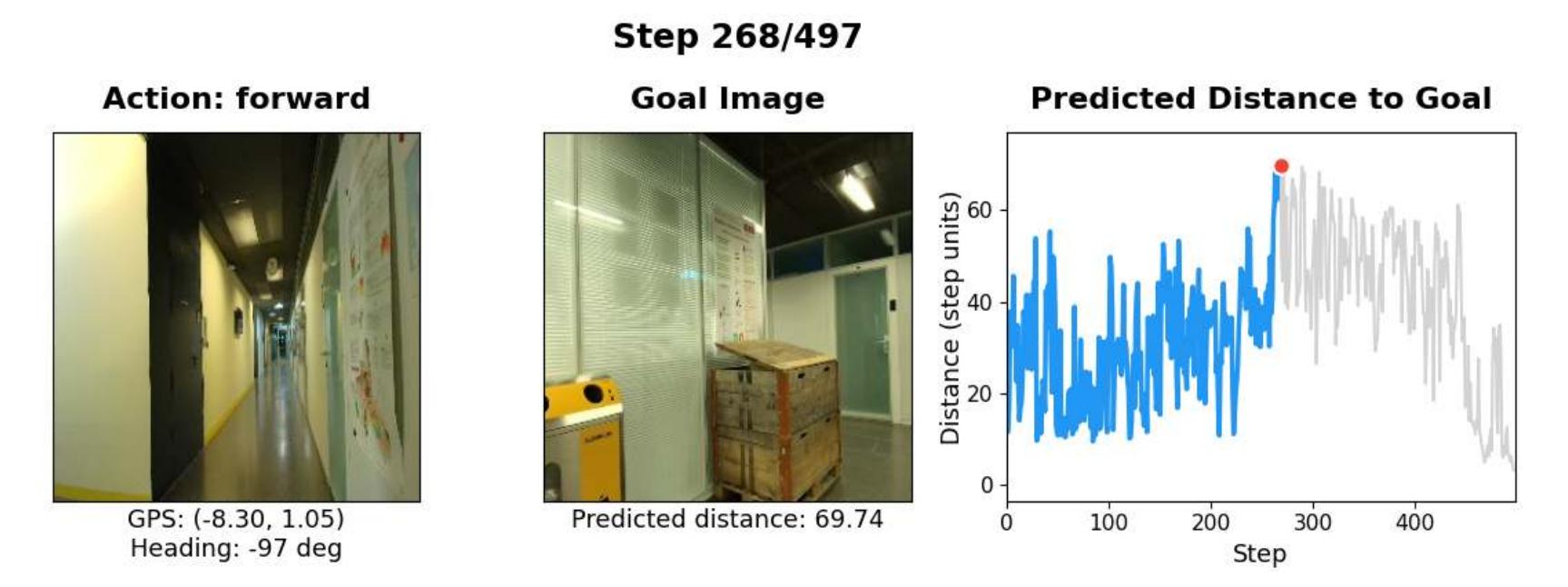}
  \caption{While the predicted distance initially increases, the agent detects a slight decrease when turning toward the left, indicating a potentially promising direction.}
\end{subfigure}

\begin{subfigure}{0.82\linewidth}
  \centering
  \includegraphics[width=\linewidth]{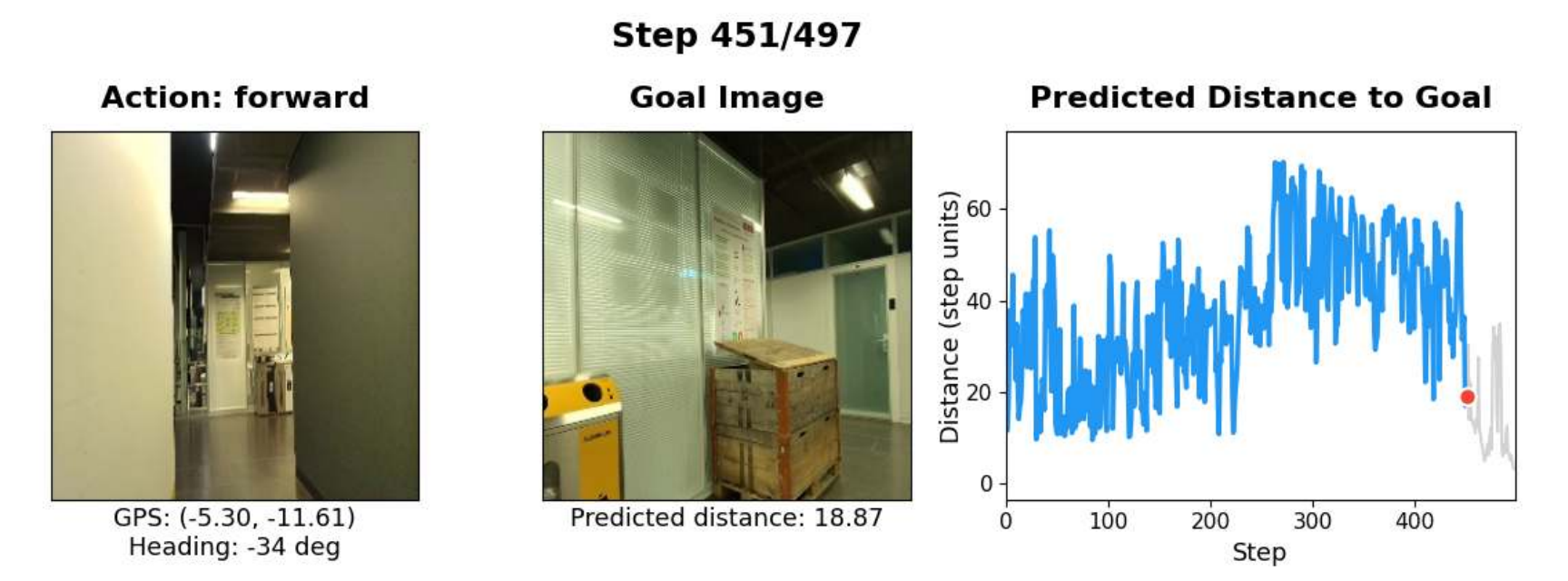}
  \caption{Moving further along the hallway reveals visual cues consistent with the goal image, causing the predicted distance to drop sharply.}
\end{subfigure}

\begin{subfigure}{0.82\linewidth}
  \centering
  \includegraphics[width=\linewidth]{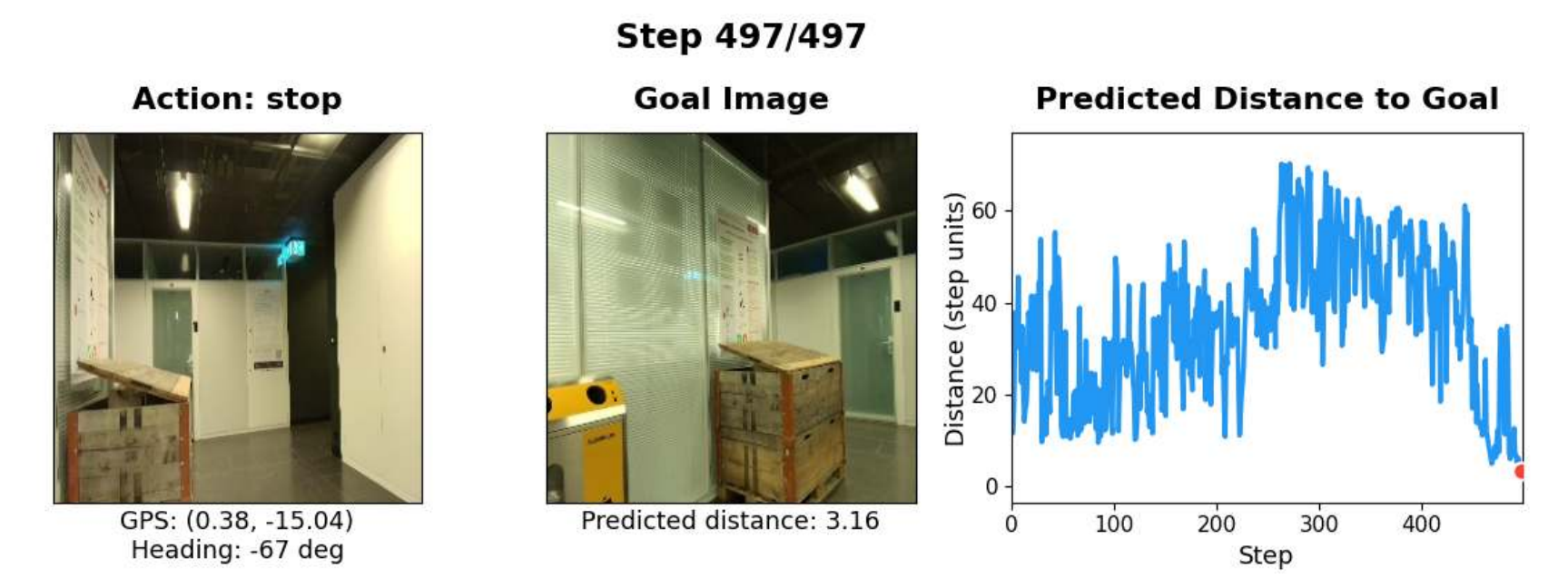}
  \caption{The agent follows the decreasing distance signal and successfully navigates to the goal.}
\end{subfigure}

\caption{\textbf{Example real-world navigation trajectory using VLD.}
Even when the goal object is not directly visible, the predicted visual distance provides directional guidance that allows the agent to progressively move toward the target.}
\label{fig:real-world-nav}
\end{figure*}

FGPrompt-EF succeeds in all five trials for $G_4$ and $G_5$. On successful trials, VLD + GeoNoise uses fewer actions on average for $G_2$ and $G_5$, while FGPrompt-EF uses fewer for $G_3$ and $G_4$ (it has no successful trial for $G_1$).
Figure~\ref{fig:real-world-nav} illustrates a representative VLD-driven trajectory.

\end{document}